%% file: ToT_TPAMI.tex
\documentclass[lettersize,journal]{IEEEtran}
\usepackage[nospace,noadjust]{cite}
\usepackage{amsmath,amssymb,amsfonts,times}
\usepackage{graphicx}
\usepackage{textcomp}
\usepackage{xcolor}
\usepackage{subfigure}
\usepackage{epsfig}
\usepackage{multirow}
\usepackage{algorithm}
\usepackage{algorithmicx}
\usepackage{algpseudocode}
\usepackage{array}
\usepackage{epstopdf}
\usepackage{color}
\usepackage{diagbox}
\usepackage{bm}

\usepackage{pdfsync}
\usepackage{todonotes}
\makeatletter
\def\@IEEEsectpunct{\ \,}
\def\paragraph{\@startsection{paragraph}{4}{\z@}{1.5ex plus 1.5ex minus 0.5ex}%
{0ex}{\normalfont\normalsize\sffamily\bfseries}}
\makeatother

\input{macro}

\begin{document}

\title{Computational and Statistical Guarantees for Tensor-on-Tensor Regression with \\ Tensor Train Decomposition}

\author{Zhen~Qin and~Zhihui~Zhu,~\IEEEmembership{Member,~IEEE}

\thanks{ZQ and ZZ are with the Department of Computer Science and Engineering, Ohio State University, Columbus, Ohio 43201, USA. (e-mail:\{qin.660,zhu.3440\}@osu.edu). This work was supported by NSF Grant CCF-2241298 and ECCS-2409701.}}




\maketitle

\begin{abstract}
Recently, a tensor-on-tensor (ToT) regression model has been proposed to generalize tensor recovery, encompassing scenarios like scalar-on-tensor regression and tensor-on-vector regression. However, the exponential growth in tensor complexity poses challenges for storage and computation in ToT regression. To overcome this hurdle, tensor decompositions have been introduced, with the tensor train (TT)-based ToT model proving efficient in practice due to reduced memory requirements, enhanced computational efficiency, and decreased sampling complexity. Despite these practical benefits, a disparity exists between theoretical analysis and real-world performance. In this paper, we delve into the theoretical and algorithmic aspects of the TT-based ToT regression model. Assuming the regression operator satisfies the restricted isometry property (RIP), we conduct an error analysis for the solution to a constrained least-squares optimization problem. This analysis includes upper error bound and minimax lower bound, revealing that such error bounds polynomially depend on the order $N+M$. To efficiently find solutions meeting such error bounds, we propose two optimization algorithms: the iterative hard thresholding (IHT) algorithm (employing gradient descent with TT-singular value decomposition (TT-SVD)) and the factorization approach using the Riemannian gradient descent (RGD) algorithm. When RIP is satisfied, spectral initialization facilitates proper initialization, and we establish the linear convergence rate of both IHT and RGD.
Notably, compared to the IHT, which optimizes the entire tensor in each iteration while maintaining the TT structure through TT-SVD and poses a challenge for storage memory in practice, the RGD optimizes factors in the so-called left-orthogonal TT format, enforcing orthonormality among most of the factors, over the Stiefel manifold, thereby reducing the storage complexity of the IHT. However, this reduction in storage memory comes at a cost: the recovery of RGD is worse than that of IHT, while the error bounds of both algorithms depend on $N+M$ polynomially.
Experimental validation substantiates the validity of our theoretical findings.
\end{abstract}


\section{Introduction}
\label{intro}
\IEEEPARstart{T}{ensor-on-tensor} (ToT) regression model \cite{lock2018tensor,raskutti2019convex,llosa2022reduced}, that extends classical regression mode to handle responses and covariates as tensors, has recently attracted increasing attention to addressing high-dimensional data in various fields such as signal processing and machine learning \cite{CichockiMagTensor15, SidiropoulosTSPTENSOR17}, chemometrics \cite{Smilde04, AcarUnsup09}, and genetic engineering \cite{HoreNature16}, as well as specific applications including latent variable models \cite{anandkumar2014tensor}, collaborative filtering \cite{bi2018multilayer}, quantum tomography \cite{qin2024quantum}, neuroimaging \cite{zhou2013tensor}, facial model refinement \cite{cai2024robust}, and the distinction of its attributes \cite{llosa2022reduced}, longitudinal relational data analysis \cite{hoff2015multilinear}, forecasting tasks \cite{liu2020low}, sparse tensor principal component analysis \cite{li2022faster}, and tensor completion \cite{ko2020fast}. Specifically, given $m$ training observations $\{(\calB_k,\calY_k)\},k = 1,\ldots,m$, the ToT regression models the relationship between the each covariate $\calB_k\in\R^{d_1\times \cdots \times d_N}$  and the corresponding multi-variate responses $\calY_k\in\R^{d_{N+1}\times \cdots \times d_{N+M}}$ as
\begin{eqnarray}
    \label{The model of ToT regression}
    \calY_k(s_{N+1},\dots,s_{N+M})  &\!\!\!\!\!\!= \!\!\!\!\!\!&\<\calB_k, \calX^\star(:,\dots,:,s_{N+1},\dots,s_{N+M})   \>\nonumber\\
     &\!\!\!\!\!\!\!\!\!\!\!\!& + \calE_k(s_{N+1},\dots,s_{N+M}),
\end{eqnarray}
where $\calX^\star\in\R^{d_1 \times \cdots \times d_{N+M}}$ denotes the regression coefficients, and $\calE_k \in \R^{d_{N+1} \times \cdots \times d_{N+M}}$ represents possible noise. This model is termed tensor-on-tensor regression because it not only handles covariates with $N$-th order features (accounting for the second word ``tensor" in the name) but also accommodates  the response as an $M$-th order tensor (accounting for the first word ``tensor" in the name). Therefore, it is also referred to as the multivariate multiple linear regression model in \cite{llosa2022reduced}. The ToT regression model encompasses various special tensor regression models, including scalar-on-tensor regression (where the response is a scalar)~\cite{Rauhut17,Han20,qin2024guaranteed}, tensor-on-vector regression (where $N=1$)~\cite{li2017parsimonious,sun2017store,guha2021bayesian}, and  scalar-on-matrix regression (i.e., the matrix sensing problem)~\cite{CandsTIT11,recht2010guaranteed,chi2019nonconvex,Zhu21TIT}.

The goal of ToT regression is to efficiently estimate the regression coefficients $\mathcal{X}^\star$ using a minimal number of training observations ${(\mathcal{B}_k, \mathcal{Y}_k)}, k=1,\dots,m$. However, as the size of $\mathcal{X}^\star$ increases exponentially, it presents challenges in terms of storage, computation, and stable recovery. Fortunately, the tensor formulation enables a concise structure for the regression coefficients through tensor decomposition, including the canonical polyadic (CP) \cite{Bro97}, Tucker \cite{Tucker66}, and tensor train (TT) \cite{Oseledets11} decompositions, which have been demonstrated effective for efficient ToT regressions \cite{liu2020low, luo2022tensor, llosa2022reduced}.
Each tensor decomposition has its advantages and drawbacks. For instance, the CP decomposition offers a storage advantage by requiring the least amount of memory, scaling linearly with $N+M$. However, determining the CP rank of a tensor is generally an NP-hard problem \cite{haastad1989tensor, de2008tensor, kolda2009tensor}. In contrast, the Tucker decomposition can be approximately computed using the higher-order singular value decomposition (HOSVD). Yet, when employing the Tucker decomposition to represent a tensor, the size of the core tensor still grows exponentially with respect to the order $N+M$. This exponential growth leads to significant memory consumption, rendering the Tucker decomposition more suitable for low-order tensors rather than high-order ones.

In comparison, the TT decomposition offers a well-balanced representation, necessitating only $O(N+M)$ parameters, while its quasi-optimal decomposition can be achieved through a sequential singular value decomposition (SVD) algorithm, commonly known as the tensor train SVD (TT-SVD) \cite{Oseledets11}. Specifically, we say $\calX\in\R^{d_1\times \cdots \times d_{N+M}}$ is in the \emph{TT format} if its $(s_1,\dots,s_{N+M})$-th element can be expressed as the following matrix product form \cite{Oseledets11}
\begin{eqnarray}
    \label{Definition of Tensor Train}
    \calX(s_1,\dots,s_{N+M})&\!\!\!\!=\!\!\!\!&\mX_1(s_1,:)\mX_2(:,s_2,:)\cdots \nonumber\\
    &\!\!\!\!\!\!\!\!&\hspace{-2cm} \mX_{N+M-1}(:,s_{N+M-1},:)\mX_{N+M}(:,s_{N+M}),
\end{eqnarray}
where matrix factors $\mX_1\in\R^{d_1\times r_1}$, $\mX_{N+M}\in\R^{r_{N+M-1}\times d_{N+M}}$ and tensor factors $\mX_i \in\R^{r_{i-1}\times d_i \times r_i}, i=2,\dots,N+M-1$. To unify the notations for the factors, we may write $\mX_1$ and $\mX_{N+M}$ as order-3 tensor with $\mX_1(1,s_1,:) = \mX_1(s_1,:)$ and $\mX_{N+M}(:,s_{N+M},1) = \mX_{N+M}(s_{N+M},:)$ and introduce $r_0 = r_{N+M} = 1$ such that $\mX_i \in\R^{r_{i-1}\times d_i \times r_i}, i=1,\dots,N+M$. Thus, the TT format can be represented by $N + M$ factors $\{{\mX}_i\}_{i\geq 1}$, with total $O((N+M)\ol d\ol r^2)$ number of parameters, where $\ol d = \max_{i=1}^{N+M} d_i$ and $\ol r = \max_{i=1}^{N+M-1} r_i$.
The dimensions $\vr = (r_1,\dots, r_{N + M-1})$ of such a  decomposition are called the \emph{TT ranks} of $\calX$ and denoted by $\text{rank}(\calX)$.\footnote{Any tensor can be decomposed in the TT format \eqref{Definition of Tensor Train} with sufficiently large TT ranks \cite[Theorem 2.1]{Oseledets11}. Indeed, there always exists a TT decomposition with $r_i \le\min\{\Pi_{j=1}^{i}d_j, \Pi_{j=i+1}^{N + M}d_j\}$ for any $i\ge 1$.} We say a TT format tensor is low-rank if $r_i$ is much smaller compared to $\min\{\Pi_{j=1}^{i}d_j, \Pi_{j=i+1}^{N+ M}d_j\}$ for most indices\footnote{When $i = 1$ or $N+ M-1$, $r_1$ or $r_{N+ M-1}$ may not be much smaller than $d_1$ or $d_{N+ M}$.} $i$ so that the total number of parameters in the factors $\{\mX_i\}$ is much smaller than the number of entries in $\calX$. We refer to any tensor for which such a low-rank TT decomposition exists as a {\it low-TT-rank} tensor. Consequently, the TT decomposition has been applied for ToT regression in \cite{liu2020low, llosa2022reduced}, demonstrating superior performance compared to other tensor decompositions. For simplicity, we refer to the ToT regression with TT decomposition as TT-based ToT regression.

Although TT-based ToT regression has achieved superior empirical performance, to the best of our knowledge, its theoretical analysis, especially regarding statistical error bounds and algorithms with guaranteed performance, has not been well studied. Only Tucker-based ToT regression has recently been theoretically studied in \cite{luo2022tensor}. In contrast, the TT-based regression has only been theoretically studied for the case where the response is a scalar. This includes the analysis of statistical robustness in the context of quantum state tomography \cite{qin2024quantum}, and algorithmic development with guaranteed convergence and performance, including algorithms that operate on the entire tensor (such as the iterative hard thresholding (IHT)\footnote{The analysis relies on an unverified perturbation bound of the TT-SVD projection and provides a recovery guarantee that may not be optimal.} \cite{Rauhut17, rauhut2015tensor} and the Riemannian gradient descent \cite{budzinskiy2021tensor, cai2022provable}) and that optimize over the factors~\cite{qin2024guaranteed}. Extending these results directly to the general ToT regression may not yield optimal conclusions. For instance, one may attempt to estimate the sub-tensors $\calX^\star(:,\dots,:,s_{N+1},\dots,s_{N+M})$ separately
for each $(s_{N+1}, \dots, s_{N+M})$ in \eqref{The model of ToT regression}. In this case, the output is a scalar, and existing results can be directly applied. However, due to $d_{N+1}\cdots d_{N+M}$ many such sub-tensors,
this approach might lead to an estimate $\wh\calX$ with an exponential large error $\| \wh\calX - \calX^\star\|_F$. This arises from disregarding the compact TT structure within the entire tensor $\calX^\star$, which is the motivation of the ToT regression models \cite{lock2018tensor,raskutti2019convex,llosa2022reduced}. Therefore, ToT regression model requires new analysis and algorithms that capture the connection between the regression coefficients $\{\calX^\star(:,\dots,:,s_{N+1},\dots,s_{N+M})\}_{(s_{N+1},\dots,s_{N+M})}$ for each response, i.e., the compact structure within the entire regression coefficients $\calX^\star$.

\begin{table*}[!ht]
\renewcommand{\arraystretch}{1.8}
\begin{center}
\caption{Comparison of statistical guarantees and minimax lower bounds for ToT regression with  Tucker and TT decompositions. Here $\gamma^2$ is the variance of each element in the noise tensor $\calE$.}
\label{Comparison among different tensor regression}
{\begin{tabular}{|c||c|c|} \hline  {Decomposition}  &{Upper bound} & {Minimax lower bound}
\\\hline {Tucker (\cite{luo2022tensor})} &  $O\bigg(\sqrt{\frac{(1+\delta_{2\wt r^\text{tk}}) (\sum_{i=1}^{N+M}r_i^\text{tk} (d_i-r_i^\text{tk}) + \Pi_{i=1}^{N+M}r_i^\text{tk}  )   }{(1-\delta_{2\wt r^\text{tk}})^2m}}\gamma\bigg)$ & $\Omega\bigg(\sqrt{\frac{\sum_{i=1}^{N+M}r_i^\text{tk} (d_i-r_i^\text{tk}) + \Pi_{i=1}^{N+M}r_i^\text{tk}}{m}}\gamma\bigg)$
\\\hline {TT} &  $O\bigg(\sqrt{\frac{(1+\delta_{2\wt r})(\sum_{i=1}^{N+M} d_i r_{i-1}r_i\log (N+M))}{(1-\delta_{2\wt r})^2m}}\gamma\bigg)$ & $\Omega\bigg(\sqrt{\frac{\sum_{j=1}^{N+M}d_jr_{j-1}r_j}{m}}\gamma\bigg) $ \\\hline
\end{tabular}}
\end{center}
\end{table*}

\paragraph*{Our contributions:}
In this paper, we aim to develop statistical analysis and optimization algorithms with guaranteed convergence for TT-based ToT regression. Before introducing our main results, we further simplify the notation by stacking all $\{\calB_k\}$, $\{\calY_k\}$, and $\{\calE_k\}$ from \eqref{The model of ToT regression} into tensors $\calB\in\R^{m\times d_1\times \cdots \times d_N}$ and $\calY, \calE\in\R^{m\times  d_{N+1} \times \cdots \times d_{N+M}}$ such that $\calB(k,:\dots,:) = \calB_k$, $\calY(k,:\dots,:) = \calY_k$ and $\calE(k,:\dots,:) = \calE_k$. We also define a linear mapping $\calA: \R^{d_{1}\times  \cdots \times d_{N + M}}\rightarrow \R^{m\times  d_{N+1} \times \cdots \times d_{N+M}}$ from the regression coefficients to the response, that is the $(k, s_{N+1},\dots, s_{N+M} )$-th element of $\calA(\calX^\star)$  is $\<\calB_k, \calX^\star(:,\dots,:,s_{N+1},\dots,s_{N+M})   \> $. Then the ToT regression model \eqref{The model of ToT regression} can be written succinctly as
\begin{eqnarray}
    \label{The compact form of ToT model}
    \calY = \calA(\calX^\star) + \calE.
\end{eqnarray}
Given the collection of covariates $\calB$ and responses $\calY$, we attempt to recover the underlying low-TT-rank regression coefficients $\calX^\star$ by solving the following TT optimization problem
\begin{eqnarray}
    \label{The loss function in ToT for TT intro}
     \argmin_{\calX\in\R^{d_1 \times \cdots \times d_{N+M}}, \atop \text{rank}(\calX) = (r_1,\dots,r_{N+M-1})} \frac{1}{2m}\|\calA(\calX) - \calY\|_F^2,
\end{eqnarray}
where again $\text{rank}(\calX)$ denotes the TT ranks of $\calX$.

\paragraph*{Statistical Guarantees} Our first main contribution concerns the global optimality of the solutions generated by the ToT model in \eqref{The loss function in ToT for TT intro}. Towards that goal, we first study the stable embeddings of TT-format tensors from linear measurements. The stable embedding result, a concept well studied in the compressive sensing literature \cite{donoho2006compressed,candes2006robust,
candes2008introduction,recht2010guaranteed,eftekhari2015new} ensures a robust recovery of the regression coefficients. Specifically, we extend the restricted isometry property (RIP) \cite{Rauhut17,qin2024quantum} from TT-based scalar-on-tensor regression to the general ToT regression and show that this condition can be satisfied with $m\gtrsim N\wt d\wt r^2\log(N\wt r)$  generic subgaussian measurements (where $\wt r=\max_{i=1}^{N-1} r_i$ and $\wt d=\max_{i=1}^N d_i$). Under the RIP condition, we show that the low-TT-rank model \eqref{The loss function in ToT for TT intro} provides a stable recovery with recovery error bounded by $O((N+M)\ol d\ol r^2\log((N+M)) \gamma^2/m)$ where $\ol r=\max_{i=1}^{N + M -1} r_i$, $\ol d=\max_{i=1}^{N + M} d_i$ and $\gamma^2$ denotes the variance of  each element in $\calE$. It is essential to highlight that when the logarithm term is neglected, the recovery bound achieves optimality compared to the degrees of freedom for the TT structure.
Formally, assuming Gaussian measurement operator $\calA$, we further establish a minimax lower bound on the estimation error for parameter estimation of TT format tensors, indicating that the obtained stable recovery result is informationally-theoretically optimal up to log factors.
The study most akin to ours is the work on the Tucker-based ToT regression model \cite{luo2022tensor}. The comparison with the Tucker-based ToT case is summarized in Table~\ref{Comparison among different tensor regression}. Here, we assume both ToT model satisfies RIP with generic subgaussian measurements: $2\wt r$-RIP for the TT case with a constant $\delta_{2\wt r}$, and $2\wt r^\text{tk}$-RIP for the Tucker case with $m\gtrsim \sum_{i=1}^{N}r_i^\text{tk} (d_i-r_i^\text{tk}) + \Pi_{i=1}^{N}r_i^\text{tk}$ and a constant $\delta_{2\wt r^\text{tk}}$, where $\wt r^\text{tk} = \max_{i=1}^{N}r_i^\text{tk} $ and $(r_1^\text{tk}, \dots, r_{N+M}^\text{tk})$ are the ranks in the Tucker decomposition \cite[Proposition 1]{luo2022tensor}. It is evident that the statistical guarantees of the TT-based ToT regression model depend only polynomially on the order $N+M$, while the ones for the Tucker-based model increase exponentially. We finally note that, compared with the expression in \Cref{Comparison among different tensor regression} for the TT case, we opt to replace the term $\sum_{i=1}^{N+M} d_i r_{i-1}r_i$ with $O((N+M)\ol d\ol r^2)$ to simplify the mathematical expression in the following presentation, particularly the main theorems.

\paragraph*{Algorithmic Guarantees} Our second main contribution is the development of efficient optimization algorithms with guaranteed performance for solving the problem in \eqref{The loss function in ToT for TT intro}. We first study the IHT algorithm, i.e., projected gradient descent, that optimizes the entire tensor in each iteration and utilizes the TT-SVD to project the iterates back to the TT format. We show that when initialized appropriately, the IHT algorithm can converge to the ground truth $\calX^\star$ and attain the aforementioned upper error bound. Notably, our convergence analysis improves upon the previous work~\cite{Rauhut17, rauhut2015tensor} for the scalar case by using a good initialization strategy along with a correspondingly tighter expansion of the TT-SVD to avoid a strong assumption on an unverified perturbation bound of the TT-SVD projection.

On the other hand, for high-order tensors, it could be prohibitive to apply the IHT algorithm or other methods that operate on the entire tensor since the tensor size increases exponentially in terms of the order. To address this issue, we propose another algorithm based on the factorization approach that directly optimizes over the factors $\{\mX_i\}$. Specifically, using $
    \calX = [\mX_1,\dots, \mX_{N+M}]$ to indicate the TT format tensor with factors $\{\mX_i\}$, we solve the following factorized optimization problem
\begin{eqnarray}
    \label{The loss function in ToT for TT factor intro}
    \begin{split}
     \min_{\{\mX_i\in\R^{r_{i-1}\times d_i\times r_i}\}_i}  & \frac{1}{2m}\|\calA([\mX_1,\dots,\mX_{N+M}]) - \calY\|_F^2.
     \end{split}
\end{eqnarray}
We extend the result in \cite{qin2024guaranteed} and show that with an appropriate initialization, a gradient-based method converges at a linear rate, with the recovery error increasing polynomially in terms of the order $N+M$, slightly worse than the IHT method. Roughly speaking, IHT operates on the entire tensor and its gradient update solves a convex objective \eqref{The loss function in ToT for TT intro} (when ignoring the projection step), generally leading to a faster convergence compared to RGD, which solves a highly nonconvex objective due to the product structure between the factors. However, IHT sacrifices storage efficiency, as it requires operating on the entire tensor at each iteration, leading to exponential memory consumption.
This highlights a trade-off between storage efficiency and accuracy in choosing the IHT and factorization approach; see \Cref{Compelxity comparison among different algorithms} in \Cref{Comparison between IHT and RGD} for a detailed comparison. In addition, we show that the commonly used spectral initialization provides a valid starting point that falls within the local basin of attraction, ensuring linear convergence for both IHT and gradient-based method for the factorization approach. We perform a series of experiments in \Cref{Sec experiment} to validate our theoretical findings.

\paragraph*{Notations} We use calligraphic letters (e.g., $\calY$) to denote tensors,  bold capital letters (e.g., $\mY$) to denote matrices, except for $\mX_i$ which denotes the $i$-th order-$3$ tensor factors in the TT format ($i=2,\dots,N+M-1$),  bold lowercase letters (e.g., $\vy$) to denote vectors, and italic letters (e.g., $y$) to denote scalar quantities.  Elements of matrices and tensors are denoted in parentheses, as in Matlab notation. For example, $\calX(s_1, s_2, s_3)$ denotes the element in position
$(s_1, s_2, s_3)$ of the order-3 tensor $\calX$.
The inner product of $\calA\in\R^{d_1\times\dots\times d_N}$ and $\calB\in\R^{d_1\times\dots\times d_N}$ can be denoted as $\<\calA, \calB \> = \sum_{s_1=1}^{d_1}\cdots \sum_{s_N=1}^{d_N} \calA(s_1,\dots,s_N)\calB(s_1,\dots,s_N) $.
$\|\calX\|_F = \sqrt{\<\calX, \calX \>}$ is the Frobenius norm of $\calX$.
$\|\mX\|$ and $\|\mX\|_F$ respectively represent the spectral norm and Frobenius norm of $\mX$.
$\sigma_{i}(\mX)$ is the $i$-th singular value of $\mX$.
For a positive integer $K$, $[K]$ denotes the set $\{1,\dots, K \}$. For two positive quantities $a,b\in \real$, the inequality $b\lesssim a$ or $b = O(a)$ means $b\leq c a$ for some universal constant $c$; likewise, $b\gtrsim a$ or $b = \Omega(a)$ indicates that $b\ge ca$ for some universal constant $c$. To simplify notations in the following sections, for an order-$(N+M)$ TT format with ranks $(r_1,\dots, r_{N+M-1})$, we define $\wt r=\max_{i=1}^{N-1} r_i$ and $\wt d=\max_{i=1}^N d_i$ in the RIP condition. In the context of statistical and computational guarantees, we define $\ol r=\max_{i=1}^{N + M -1} r_i$ and $\ol d=\max_{i=1}^{N + M} d_i$.

\section{Statistical Guarantees for TT-ToT Regression}
\label{Tensor-on-tensor Regression}

\subsection{Tensor Train Decomposition}

We begin by introducing some useful results on TT decomposition.
Recall the TT format in \eqref{Definition of Tensor Train}.
Since $\mX_1(s_1,:)$, $\mX_{N+M}(:,s_{N+M})$ and $\mX_{i}(:,s_i,:),i=2,\dots,N+M-1$ will be extensively used, we will denote them simply by  ${\mX_i(s_i)}\in\R^{r_{i-1}\times r_{i}}$. The $(s_1,\dots,s_{N+M})$-th element in $\calX$ can then be written as $\calX(s_1,\dots,s_{N+M})=\prod_{i=1}^{N+M}{\mX_i(s_i)}$.
We define the set of TT format tensors with maximum TT rank equal to $\ol r =\max_i r_i$:
\begin{eqnarray}
    \label{The set of TT}
    \setX_{\ol{r}} &\!\!\!\!=\!\!\!\!& \{\calX \in\R^{d_1\times\cdots\times d_{N+M}}: \text{rank}(\calX) = (r_{1},\dots,r_{N+M-1}), \nonumber\\
    &\!\!\!\!\!\!\!\!&\ol r=\max_i r_i,  \}.
\end{eqnarray}

The decomposition of the tensor $\calX$ into the form of \eqref{The set of TT} is generally not unique: not only the factors ${\mX_i(s_i)}$  are not unique, but also the dimension of these factors can vary. We say the decomposition \eqref{left orthogonal form L} is \emph{minimal} if the left unfolding matrix $L(\mX_i)$, defined as
\begin{eqnarray}
    \label{left orthogonal form L}
    L(\mX_i)=\begin{bmatrix}\mX_i(1) \\ \vdots\\  \mX_i(d_i) \end{bmatrix}\in\R^{(r_{i-1}d_i) \times r_i},
\end{eqnarray}
has full column rank, i.e., the rank of $L(\mX_i)$ is $r_i$. The dimensions $\vr = (r_1,\dots, r_{N + M -1})$ of such a minimal decomposition are called the \emph{TT ranks} of $\calX$, while the largest value $\ol r= \max_i r_i$ may also be referred to as the \emph{TT rank}.
According to \cite{holtz2012manifolds}, there is exactly one set of ranks $\vr$ that $\calX$ admits a minimal TT decomposition. In this case, $r_i$ also equals to the rank of the $i$-th unfolding matrix $\calX^{\<i\>}\in\R^{(d_1\cdots d_i)\times (d_{i+1}\cdots d_{N+M})}$ of the tensor $\calX$, where the $(s_1\cdots s_i, s_{i+1}\cdots s_{N+M})$-th element\footnote{ Specifically, $s_1\cdots s_i$ and $s_{i+1}\cdots s_{N+M}$ respectively represent the $(s_1+d_1(s_2-1)+\cdots+d_1\cdots d_{i-1}(s_i-1))$-th row and $(s_{i+1}+d_{i+1}(s_{i+2}-1)+\cdots+d_{i+1}\cdots d_{N+M-1}(s_{N+M}-1))$-th column.
} of $\calX^{\<i\>}$ is given by $\calX^{\<i\>}(s_1\cdots s_i, s_{i+1}\cdots s_{N+M}) = \calX(s_1,\dots, s_{N+M})$.
This can also serve as an alternative way to define the TT ranks.
With the $i$-th unfolding matrix $\calX^{\<i\>}$\footnote{We can also define the $i$-th unfolding matrix as $\calX^{\< i \>} = \mX^{\leq i}\mX^{\geq i+1}$, where each row of the left part $\mX^{\leq i}$ and each column of the right part $\mX^{\geq i+1}$ can be represented as $\mX^{\leq i}(s_1\cdots s_i,:) = \mX_1(s_1)\cdots \mX_i(s_i)$ and $\mX^{\geq i+1}(:,s_{i+1}\cdots s_{N+M}) = \mX_{i+1}(s_{i+1})\cdots \mX_{N+M}(s_{N+M})$. When factors are in left-orthogonal form satisfying \eqref{orthogonal property of left orthogonal}, we have ${\mX^{\leq i}}^\top\mX^{\leq i} = \mId_{r_i}$ and $\sigma_j(\calX^{\< i\>}) = \sigma_j( \mX^{\geq i+1} ), j\in[N+M-1]$.} and \emph{TT ranks}, we can obtain its smallest singular value $\underline{\sigma}(\calX)=\min_{i=1}^{N + M -1}\sigma_{r_i}(\calX^{\<i\>})$, its largest singular value $\overline{\sigma}(\calX)=\max_{i=1}^{N + M-1}\sigma_{1}(\calX^{\<i\>})$ and condition number $\kappa(\calX)=\frac{\overline{\sigma}(\calX)}{\underline{\sigma}(\calX)}$.

Moreover, for any TT format $\calX$ of form \eqref{The set of TT} with a minimal decomposition, there always exists a factorization such that $L(\mX_i)$ are orthonormal matrices for all $i \in [N + M -1]$; that is \cite{holtz2012manifolds}
\begin{eqnarray}
\label{orthogonal property of left orthogonal}
    L^\top(\mX_i)L(\mX_i) = \mId_{r_i}, \ \forall i=1,\dots,N + M -1.
\end{eqnarray}
The resulting TT decomposition is called the {\it left-orthogonal form}, or {\it left-canonical form}.
Since any tensor can be represented in the left-orthogonal TT form, the set $\setX_{\ol{r}}$ in \eqref{The set of TT} is equivalent to one space encompassing the entirety of left-orthogonal TT formats. For the sake of simplicity, we assume that all tensors in this paper belong to $\setX_{\ol{r}}$. Should there arise a necessity to transform a tensor into the left-orthogonal form, we will provide additional elucidation.

\subsection{Statistical Guarantees for TT-ToT Regression}
In the TT-based ToT regression model, our objective is to retrieve the ground-truth coefficients $\calX^\star\in\setX_{\ol{r}}$ from the training covariates $\calB$ and responses $\calY$ as in \eqref{The compact form of ToT model} by minimizing the following constrained least squares objective:
\begin{eqnarray}
    \label{The loss function in ToT for TT}
    \wh\calX = \argmin_{\calX\in\setX_{\ol{r}}} \frac{1}{2m}\|\calA(\calX) - \calY\|_F^2,
\end{eqnarray}
where $\calA$ is a linear map defined in \eqref{The compact form of ToT model} with $(k, s_{N+1}, \dots,\\ s_{N+M} )$-th element $\calA(\calX)(k, s_{N+1},\dots, s_{N+M} )$ being computed via $\<\calB(k,:\dots,:), \calX(:,\dots,:,s_{N+1},\dots,s_{N+M})   \> $.
To facilitate the retrieval of the tensor $\calX^\star$ from its training observations, the covariates should adhere to specific properties. A particularly valuable characteristic is the Restricted Isometry Property (RIP), which is widely applied in the tensor regression \cite{grotheer2021iterative, Rauhut17, qin2024quantum}.
By using standard covering argument, we can obtain the following TT-based tensor-on-tensor RIP.
\begin{theorem}
\label{RIP condition fro the ToT TT regression Lemma}
Suppose all elements in covariates $\calB$, constituting the linear map $\calA: \R^{d_{1}\times  \cdots \times d_{N+M}}\rightarrow \R^{m\times d_{N+1}\times  \cdots \times d_{N+M}}$, are independent subgaussian random variables with mean zero and variance one, such as the Gaussian random variable and the Bernoulli random variable.
Let $\delta_{\wt r}\in(0,1)$ be a positive constant. Then, for any TT format $\calX\in\R^{d_{1} \times \cdots \times d_{N+M}}$ with the rank $(r_1,\dots, r_{N+M-1})$, when the number of covariates satisfies
\begin{equation}
m \ge C \cdot \frac{1}{\delta_{\wt r}^2} \cdot \max\left\{ N\wt d{\wt r}^2\log(N\wt r), \log(1/\epsilon)\right \},
\label{eq:mrip ToT TT}
\end{equation}
with probability at least $1-\epsilon$, $\calA$ satisfies the $\wt r$-RIP:
\begin{eqnarray}
    \label{RIP condition fro the ToT TT regression}
    (1-\delta_{\wt r})\|\calX\|_F^2\leq \frac{1}{m}\|\mathcal{A}(\calX)\|_F^2\leq(1+\delta_{\wt r})\|\calX\|_F^2,
\end{eqnarray}
where $C$ is a universal constant depending only on $L$.
\end{theorem}
The proof is given in \Cref{Proof of ToT RIP} of the supplementary material.
\Cref{RIP condition fro the ToT TT regression Lemma} ensures  a stable embedding for TT format tensors and guarantees that the energy $\|\mathcal{A}(\calX)\|_F^2$ is close to $\|\calX\|_F^2$. For any two distinct TT format tensors $\calX_1,\calX_2$ with the first $N-1$ TT ranks smaller than $\wt r$, noting that $\calX_1 - \calX_2$ is also a TT format tensor with the first $N-1$ TT ranks smaller than $2\wt r$ according to \cite[eq. (3)]{qin2024guaranteed}, the RIP condition ensures distinct responses $\calA(\calX_1)$ and $\calA(\calX_2)$ as
\begin{eqnarray}
\frac{1}{m}\|\calA(\calX_1)- \calA(\calX_2)\|_2^2 &\!\!\!\!=\!\!\!\!& \frac{1}{m}\|\calA(\calX_1- \calX_2)\|_2^2\nonumber\\
&\!\!\!\! \ge \!\!\!\! & (1- \delta_{2\wt r}) \|\calX_1- \calX_2\|_F^2,
\end{eqnarray}
which guarantees the possibility of exact recovery.
Moreover, it is important to observe that the number of training observations $m$ scales linearly with the tensor order $N$ rather than $N+M$, given that the size of $\calA(\calX)$ is $m\times d_{N+1}\times \cdots\times d_{N+M}$. In contrast to the case with scale responses, the inclusion of dimensions $d_{N+1}\times\cdots\times d_{N+M}$ in the response facilitates precise recovery, consequently reducing the required number of measurement operators $m$ to ensure the RIP property.

We now present a formal analysis for the recovery performance of $\wh \calX$. Using the facts that $\wh \calX$ is a global minimum of \eqref{The loss function in ToT for TT} and that $\calX^\star\in\setX_{\ol r}$, we have
\begin{eqnarray}
    \label{whX and X star relationship}
    0 &\!\!\!\!\leq\!\!\!\!& \frac{1}{m}\|\calA(\calX^\star) - \calY\|_F^2  - \frac{1}{m}\|\calA({\wh\calX}) - \calY\|_F^2\nonumber\\
    &\!\!\!\!=\!\!\!\!& \frac{1}{m}\|\calA(\calX^\star)-\calA(\calX^\star) - \calE\|_F^2\nonumber\\
    &\!\!\!\!\!\!\!\!&- \frac{1}{m}\|\calA(\wh\calX)-\calA(\calX^\star) - \calE\|_F^2\nonumber\\
    &\!\!\!\!=\!\!\!\!& \frac{2}{m}\<\calA(\calX^\star)+\calE, \calA(\wh\calX - \calX^\star)  \> + \frac{1}{m}\|\calA(\calX^\star)\|_F^2\nonumber\\
    &\!\!\!\!\!\!\!\!&- \frac{1}{m}\|\calA(\wh\calX)\|_F^2\nonumber\\
    &\!\!\!\!=\!\!\!\!& \frac{2}{m}\<  \calE, \calA(\wh\calX - \calX^\star) \> - \frac{1}{m}\|\calA(\wh\calX  -  \calX^\star)\|_F^2,
\end{eqnarray}
which further implies that
\begin{eqnarray}
    \label{wX and X star relationship_1}
    \frac{1}{m}\|\calA(\wh\calX  -  \calX^\star)\|_F^2 \leq \frac{2}{m}\<  \calE, \calA(\wh\calX - \calX^\star) \>.
\end{eqnarray}

By \Cref{RIP condition fro the ToT TT regression Lemma}, we can establish the lower bound for the left-hand side of \eqref{wX and X star relationship_1}: $\frac{1}{m}\|\calA(\wh\calX  -  \calX^\star)\|_F^2 \geq (1-\delta_{2\wt r})\|\wh\calX  -  \calX^\star\|_F^2$. For the right hand side of \eqref{wX and X star relationship_1}, a direct Cauchy-Schwarz inequality $\<  \calE, \calA(\wh\calX - \calX^\star) \> \le \|\calE\|_F\|\calA(\wh\calX - \calX^\star)\|_F$ could lead to suboptimal results since the term $\|\calE\|_F$ could be exponentially large. Instead, we will exploit the statistical property of the random noise $\calE$ (each element follows a normal distribution $\calN(0,\gamma^2)$) and use again a covering argument to establish a tight bound, which leads to following main result. We defer the detailed analysis to \Cref{Proof of upper bound for error difference} of the supplementary material.

\begin{theorem} (Upper bound of $\|\wh \calX - \calX^\star\|_F$)
\label{Upper bound of error difference} Suppose that $\calA$ satisfies $2\wt r$-RIP for any tensor $\calX^\star$ in \eqref{The set of TT}, and that each element in  $\calE\in\R^{m\times d_{N+1}\times\cdots\times d_{N+M}}$ follows a normal distribution $\calN(0,\gamma^2)$. Then with probability at least $1-e^{-\Omega((N+M)\ol d\ol r^2 \log (N+M))}$, the solution $\wh \calX$ of \eqref{The loss function in ToT for TT} satisfies
\begin{align}
    \label{upper bound of error final_ conclusion}
    \|\wh \calX - \calX^\star\|_F= O\bigg(\frac{\ol r\gamma\sqrt{(1+\delta_{2\wt r})(N+M)\ol d \log (N+M)}}{(1-\delta_{2\wt r})\sqrt{m}}\bigg).
\end{align}
\end{theorem}

We first note that as the noise impacts all the $N+M$ factors of the tenor $\calX^\star$, the recovery guarantee at the right-hand side of \eqref{upper bound of error final_ conclusion} has term $\sqrt{(N+M)\ol d}\ol r$ in the numerator that depends on the entire $N+M$ factors; here $\ol r=\max_{i=1}^{N + M -1} r_i$ and $\ol d=\max_{i=1}^{N + M} d_i$. \Cref{Upper bound of error difference} guarantees a stable recovery of the ground truth with recovery error $O(\ol r\gamma\sqrt{(N+M)\ol d/m})$ (ignoring the log term). In particular, thanks to the compact structure inherent in the TT format, the upper bound only scales linearly (except for the log terms) with respect to the order $N+M$, which is superior to the exponential dependence by the Tucker model \cite{luo2022tensor}.
Our following result, minimax lower bound for the TT-based ToT regression model in \eqref{The compact form of ToT model}, reveals that the upper bound is tight and optimal up to the log terms.

\begin{theorem}(Minimax lower bound of $\|\wh{\calX} - \calX\|_F$)
\label{minimax bound of error difference}
In the context of the TT-based ToT regression model in \eqref{The compact form of ToT model}, we consider $\calX \in \setX_{\ol{r}}$ in \eqref{The set of TT}.
Suppose $\min_{i=1}^{N+M-1} r_i\geq C$ for some absolute constant $C$. When each element of $\calB$ and $\calE$ in \eqref{The compact form of ToT model} respectively follow $\calN(0,1)$ and $\calN(0,\gamma^2)$, we get
\begin{eqnarray}
    \label{minimax bound of error final_ conclusion}
    \inf_{\wh \calX}\sup_{\calX\in\setX_{\ol{r}}}\E{\|\wh{\calX} - \calX\|_F}= \Omega\bigg(\sqrt{\frac{\sum_{j=1}^{N+M}d_jr_{j-1}r_j}{m}}\gamma\bigg).
\end{eqnarray}
\end{theorem}
The detailed proof has been provided in \Cref{proof of minimax bound for TT}  of the supplementary material.
In the following section, we will propose two algorithms that achieve the minimax lower bound.

\section{Optimization Methods for TT-ToT Regression}
\label{Sec optimization method}
In this section, we present two gradient-based optimization algorithms. We first introduce an iterative hard thresholding (IHT) algorithm, which projects the estimated tensor onto $\setX_{\ol{r}}$ via the TT-SVD after a gradient-based update in each iteration. This method can converge to a point that achieves recovery guarantee as in \Cref{Upper bound of error difference}, but at the cost of large storage memory as it operates on the entire tensor in each iteration. To solve the problem of high-order tensors which could be exponentially large, we propose another algorithm based on the factorization approach that directly optimizes over the factors. This approach reduces storage memory but comes with a slightly worse upper error bound than the IHT. Finally, we show that the commonly used spectral initialization provides a valid starting point that falls within the local basin of attraction, ensuring linear convergence for both IHT and gradient-based method for the factorization approach.

\subsection{Iterative hard thresholding algorithm}
\label{IHT method}
We start by rephrasing the loss function in \eqref{The loss function in ToT for TT} as follows:
\begin{eqnarray}
    \label{The loss function in ToT for TT f(X)}
     \min_{\calX\in\setX_{\ol{r}}} g(\calX) = \frac{1}{2m}\|\calA(\calX) - \calY\|_F^2.
\end{eqnarray}
Since this is a minimization problem over the set $\setX_{\ol{r}}$ of low-TT-rank tensors, we can utilize the projected gradient descent method, also known as the iterative hard thresholding (IHT) algorithm in the compressive sensing literature \cite{Rauhut17}. Specifically, in each iteration, IHT performs a gradient descent update, followed by a projection onto the set $\setX_{\ol{r}}$. While computing the optimal projection onto the set $\setX_{\ol{r}}$ could be NP-hard, a sub-optimal projection could be obtained by TT-SVD \cite{Oseledets11}, denoted by $\text{SVD}_{\vr}^{tt}(\cdot)$ here. Thus, we use the following IHT algorithm:
\begin{eqnarray}
    \label{Iterative equ of l2 GDwithTT_SVD_1}
    \calX^{(t+1)} = \text{SVD}_{\vr}^{tt}(\calX^{(t)} - \mu\nabla g(\calX^{(t)})),
\end{eqnarray}
where the explicit form of the gradient $\nabla g(\calX^{(t)})$ is defined in \Cref{proof of converegence L2 IHT} of the supplementary material (see \eqref{The definition of gradient in the IHT TOT}). The following theorem establishes a local convergence performance of IHT for tensor-on-tensor regression.

\begin{theorem}
\label{Convergence analysis of IHT L2}
In the ToT regression model \eqref{The compact form of ToT model}, assume that $\calX^\star$ is in TT format  with ranks $\vr = (r_1,\dots, r_{N + M-1})$, that $\calA$ satisfies $4 \wt r$-RIP, and that each element in  $\calE$ follows the normal distribution $\calN(0,\gamma^2)$. Also suppose that IHT in \eqref{Iterative equ of l2 GDwithTT_SVD_1} starts with $\calX^{(0)}$ satisfying
\begin{eqnarray}
\label{initialization requirement L2 SVD}
\|\calX^{(0)} - \calX^\star\|_F\leq \frac{(1 - \delta_{4\wt r})^2\underline{\sigma}({\calX^\star})}{600(N+M) (1 + \delta_{4\wt r}^2 + 6 \delta_{4\wt r})},
\end{eqnarray}
and uses step size $\frac{\frac{600(N+M)}{\underline{\sigma}({\calX^\star})}\|\calX^{(0)} - \calX^\star \|_F}{(1+\frac{600(N+M)}{\underline{\sigma}({\calX^\star})}\|\calX^{(0)} - \calX^\star \|_F)(1-\delta_{4\wt r})}<\mu \leq \frac{1 - \delta_{4\wt r}}{2(1 + \delta_{4\wt r})^2}$.

Then with probability at least $1-e^{-\Omega((N+M)\ol d\ol r^2 \log (N+M))}$, we can guarantee that
\begin{eqnarray}
    \label{The expansion of difference two tensors in l2 conclusion theorem}
    \| \calX^{(t+1)}   - \calX^\star \|_F^2 &\!\!\!\!\leq\!\!\!\!& a^{t+1}\|\calX^{(0)} - \calX^\star\|_F^2\nonumber\\
    &\!\!\!\!\!\!\!\!& + O(b\|\calX^{(0)} - \calX^\star\|_F + b^2 ),
\end{eqnarray}
where $a = \big( 1+ \frac{600(N+M)}{\underline{\sigma}({\calX^\star})}\|\calX^{(0)} - \calX^\star \|_F \big)(1 - \mu(1 - \delta_{4\wt r})   ) < 1$ and $b = O(\frac{\ol r\gamma\sqrt{(1+\delta_{2\wt r})(N+M)\ol d(\log (N+M))}}{\sqrt{m}})$.
\end{theorem}
The detailed proof is given in \Cref{proof of converegence L2 IHT} of the supplementary material. First note that when $\mu$ is appropriately chosen, the convergence rate $a$ is less than 1, indicating a linear convergence of IHT. Furthermore, the upper error bound at the right hand side of \eqref{The expansion of difference two tensors in l2 conclusion theorem} is dominated by $b^2$, aligning with the error bound in \eqref{upper bound of error final_ conclusion}.
In contrast to the analysis presented in \cite{Rauhut17} for IHT that relies on an unverified perturbation bound for TT-SVD, our result exploits the guarantee of TT-SVD in \cite{cai2022provable}  and has no unproven assumptions on TT-SVD. The only requirement on the initialization will be validated in \Cref{spectral initilazation}.

\begin{table*}[!ht]
\renewcommand{\arraystretch}{1.4}
\begin{center}
\caption{Complexity comparison between IHT and RGD.}
\label{Compelxity comparison among different algorithms}
{\begin{tabular}{|c||c|c|}\hline  {Algorithm} & {Space Complexity} &{Computational Complexity}
\\\hline\hline {IHT} & $O(m\ol d^{N} + \ol d^{N+M})$ &  $O(m(\ol d^{N} + \ol d^{N+M}) + \ol d^{N+M}\ol r^2)$
\\\hline {RGD} & $O(m\ol d^{N} + (N+M)\ol d \ol r^2)$ &  $O(m(\ol d^{N} + \ol d^{N+M}) + (N+M)\ol d^{N+M}\ol r^2 + (N+M) \ol d\ol r^3)$ \\\hline
\end{tabular}}{}
\end{center}
\end{table*}

\subsection{Factorization approach---Riemannian gradient descent}
\label{RGD method}

While the IHT algorithm has demonstrated favorable performance in terms of convergence rate and error bound, its application to the high-order tensor can be prohibitive due to the exponential increase in tensor size with respect to the order. An alternative and more scalable approach is to directly optimize over the factors instead of the entire tensor, as in \eqref{The loss function in ToT for TT factor intro}. This factorization approach has been employed for solving problems such as SDPs \cite{burer2003nonlinear}, matrix sensing and completion \cite{sun2016guaranteed,Zhu18TSP,chi2018nonconvex}, and tensor PCA \cite{Han20}, tensor sensing and completion \cite{xia2019polynomial,tong2022scaling}. For the TT-based optimization problem, the work \cite{qin2024guaranteed} provides the first analysis of the factorization approach for solving the scalar-on-tensor regression, a special case of the ToT regression \eqref{The model of ToT regression}.  As mentioned in the introduction, directly applying the results from scalar responses to tensor responses might result in an exponentially large recovery error. Here, we extend the analysis in \cite{qin2024guaranteed} to the general ToT regression and study how the convergence depends on the RIP, the TT structure, and the noise level.

One of the main challenges in the analysis of iterative algorithms for the factorization approach \eqref{The loss function in ToT for TT factor intro} lies in the form of products among multiple matrices in \eqref{Definition of Tensor Train}:  the factorization is not unique, and there exist infinitely many equivalent factorizations since for any factorization $\{{\mX}_1,\dots,{\mX}_{N+M}\}$ of $\calX$,  $\{{\mX}_1 \mP_1, \mP_1^{-1}\mX_2 \mP_2,\ldots,\mP_{N+M-1}^{-1}\mX_{N+M}\}$ is also a TT factorization of $\calX$ with any invertible matrices $\mP_i\in\R^{r_i\times r_i}, i\in[N + M -1]$, $\mP_{i-1}^{-1}\mX_i \mP_i$ refers to $\mP_{i-1}^{-1}\mX_i(:,s_i,:) \mP_i$ for all $s_i\in[d_i]$. This implies that the factors could be unbalanced (e.g., $\mP_i = t\mId$ with either very large or small $t$), which makes the convergence analysis difficult. To address the ambiguity issue in the factorization approach, as in \cite{qin2024guaranteed}, we restrict all of the factors except the last one to be orthonormal and consider optimizing over the left-canonical form as follows:
\begin{eqnarray}
    \label{The loss function in ToT for TT factor}
    \begin{split}
     \min_{\mX_i\in\R^{r_{i-1}\times d_i\times r_i}} & f(\mX_1,\dots,\mX_{N+M})\\
   =& \frac{1}{2m}\|\calA([\mX_1,\dots,\mX_{N+M}]) - \calY\|_F^2,\\
\st \ & L^\top({\mX}_i)L({\mX}_i)=\mId_{r_i},i \in [N + M-1].
     \end{split}
\end{eqnarray}
When viewing each $L({\mX}_i),i \in [N + M-1]$ as a point on the Stiefel manifold which encompasses all left orthonormal matrices, we can utilize the Riemannian gradient descent (RGD) on the Stiefel manifold to optimize \eqref{The loss function in ToT for TT factor}, giving the following update
{\small \begin{align}
    \label{RIEMANNIAN_GRADIENT_DESCENT ToT_1_1}
    L(\mX_i^{(t+1)})&=\text{Retr}_{L(\mX_i)}\bigg(L(\mX_i^{(t)})-\frac{\mu}{\ol{\sigma}^2(\calX^\star)}\calP_{\text{T}_{L({\mX_i})} \text{St}}\big(\nonumber\\
    &\nabla_{L({\mX}_{i})}f(\mX_1^{(t)}, \dots, \mX_{N+ M}^{(t)})\big) \bigg), \ \ i\in [N + M -1],\\
    \label{RIEMANNIAN_GRADIENT_DESCENT ToT_1_2}
    L(\mX_{N+M}^{(t+1)}) &= L(\mX_{N+M}^{(t)})-\mu \nabla_{L({\mX}_{N+M})}f(\mX_1^{(t)}, \dots, \mX_{N+M}^{(t)}),
\end{align}}
where the expressions for the gradients are given in \Cref{proof of ToT factorization} of the supplementary material (see \eqref{the defi of gradient}). Here $\calP_{\text{T}_{L({\mX_i})} \text{St}}(\cdot)$ denotes the projection onto the tangent space of the Stifel manifold at the point $L({\mX_i})$ and $\text{Retr}_{L(\mX_i)}$ denotes the retraction operator:
{\small \begin{eqnarray}
    \label{Projection of gradient to Riemannian one}
    \hspace{-0.3cm}\calP_{\text{T}_{L({\mX_i})} \text{St}}\big(\nabla_{L({\mX}_{i})}f\big) &\!\!\!\!\!\!=\!\!\!\!\!\!& \nabla_{L({\mX}_{i})}f -\frac{1}{2}L(\mX_i^{(t)}) \nonumber\\
    &\!\!\!\!\!\!\!\! \!\!\!\!& \hspace{-2.2cm} \cdot\big((\nabla_{L({\mX}_{i})}f)^\top L(\mX_i^{(t)})+(L(\mX_i^{(t)}))^\top\nabla_{L({\mX}_{i})}f\big),\nonumber
\end{eqnarray}
\begin{eqnarray}
    \label{The definition of retraction}
    \hspace{ 0.2cm}\text{Retr}_{L(\mX_i)}(\wt L(\mX_i^{(t+1)})) = \wt L(\mX_i^{(t+1)})((\wt L(\mX_i^{(t+1)}))^\top \wt L(\mX_i^{(t+1)}))^{-\frac{1}{2}},\nonumber
\end{eqnarray}}
where we simplify $\nabla_{L({\mX}_{i})}f(\mX_1^{(t)}, \dots, \mX_{N+ M}^{(t)})$ as $\nabla_{L({\mX}_{i})}f$.

The following result ensures that, provided an appropriate initialization, the above RGD can converge linearly to a specific distance proportional to the noise variance.
\begin{theorem}
\label{Local Convergence of Riemannian in the noisy ToT regression_Theorem calX}
Suppose that $\calX^\star$ is in the TT format with ranks $\vr = (r_1,\dots, r_{N + M-1})$. Assume that $\calA$ obeys the $(N + M +3)\wt r$-RIP with a constant $\delta_{(N + M +3)\wt r}\leq\frac{7}{30}$. Suppose that the RGD in \eqref{RIEMANNIAN_GRADIENT_DESCENT ToT_1_1} and \eqref{RIEMANNIAN_GRADIENT_DESCENT ToT_1_2} is initialized with $\calX^{(0)}$ satisfying
\begin{eqnarray}
    \label{Local Convergence of Riemannian in the noisy ToT regression_Theorem initialization calX}
    \|\calX^{(0)} - \calX^\star \|_F^2\leq O\bigg(\frac{\underline{\sigma}^4(\calX^\star)}{(N+M)^4 {\ol r}^2\ol{\sigma}^2(\calX^\star)}\bigg)
\end{eqnarray}
and uses the step size $\mu=\frac{7 - 30\delta_{(N + M +3)\wt r}}{20(9(N+M)-5)(1+\delta_{(N + M +3)\wt r})^2}$.  Then, with probability at least $1 - e^{-\Omega((N+M)\ol d\ol r^2 \log (N+M))}$, the iterates $\{\calX^{(t)} \}_{t\geq 0}$ generated by the RGD satisfy
\begin{eqnarray}
    \label{Local Convergence of Riemannian in the noisy ToT regression_Theorem_1 calX}
    &\!\!\!\!\!\!\!\!&\|\calX^{(t+1)} - \calX^\star \|_F^2 \nonumber\\
    &\!\!\!\!\leq\!\!\!\! &\bigg( 1-O(\frac{1}{(N+M)^2\ol r\kappa^2(\calX^\star)}) \bigg)^{t+1}\cdot O\bigg(\frac{\underline{\sigma}^2(\calX^\star)}{(N+M)^2\ol r^2}\bigg) \nonumber\\
    &\!\!\!\!\!\!\!\!&+ O\bigg( \frac{(N+M)^5\ol d\ol r^3(\log (N+M)) \kappa^2(\calX^\star)\gamma^2}{m}\bigg),
\end{eqnarray}
as long as $m\geq C \frac{(N+M)^5\ol d \ol r^3 (\log (N+M))\ol{\sigma}^2(\calX^\star) \gamma^2}{\underline{\sigma}^4(\calX^\star)} $ with a universal constant $C$.
\end{theorem}
The proof is given in \Cref{proof of ToT factorization} of the supplementary material. We have used a restricted step size to simplify the expression for \eqref{Local Convergence of Riemannian in the noisy ToT regression_Theorem_1 calX}, but we note that the results hold as long as the step size is not too large, particularly  $\mu\leq\frac{7 - 30\delta_{(N + M +3)\wt r}}{20(9(N+M)-5)(1+\delta_{(N + M +3)\wt r})^2}$; see  \Cref{Local Convergence of Riemannian in the noisy ToT regression_Theorem} in  \Cref{proof of ToT factorization} of the supplementary material for the details. Moreover, our findings illustrate that when the initial condition $O(\frac{\underline{\sigma}^4(\calX^\star)}{(N+M)^4 {\ol r}^2\ol{\sigma}^2(\calX^\star)})$ is satisfied,  the RGD exhibits a linear convergence rate of $1-O(\frac{1}{(N+M)^2\ol r\kappa^2(\calX^\star)})$ and converges to the region $\|\calX^{(t+1)} - \calX^\star\|_F^2\leq O( \frac{(N+M)^5\ol d\ol r^3(\log (N+M)) \ol{\sigma}^2(\calX^\star)\gamma^2}{m\underline{\sigma}^4(\calX^\star)})$. These results highlight that the decay in the initial requirement, the convergence rate, and recovery error depend on $N+M$ polynomially, rather than exponentially. Compared with the results presented in \Cref{Convergence analysis of IHT L2}, RGD exhibits a less favorable initial requirement and upper error bound, albeit with the advantage of reduced storage memory usage.

\begin{figure*}[htbp]
\centering
\subfigure[]{
\begin{minipage}[t]{0.3\textwidth}
\centering
\includegraphics[width=5.5cm]{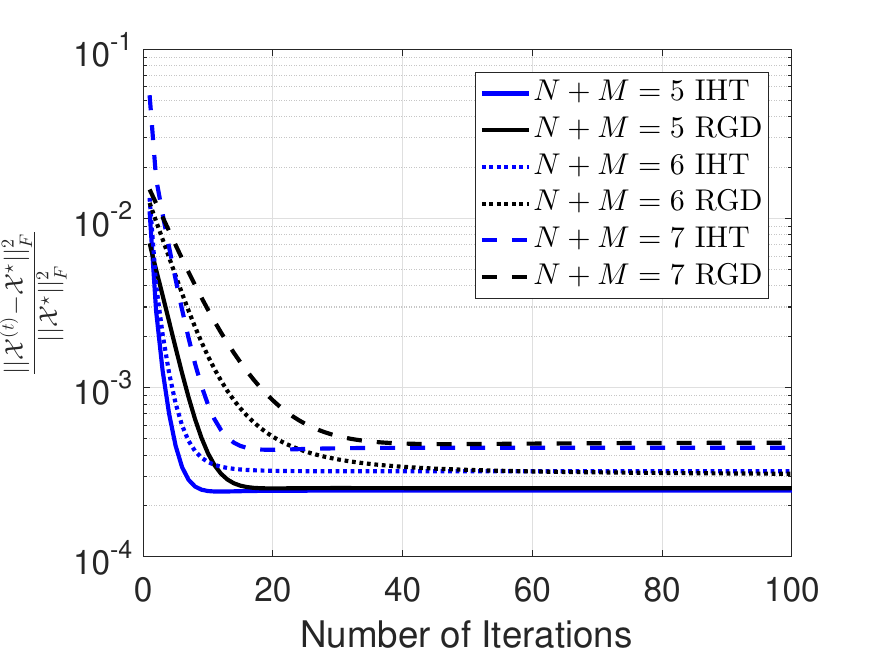}
\end{minipage}
\label{TT_ToT_N}
}
\subfigure[]{
\begin{minipage}[t]{0.3\textwidth}
\centering
\includegraphics[width=5.5cm]{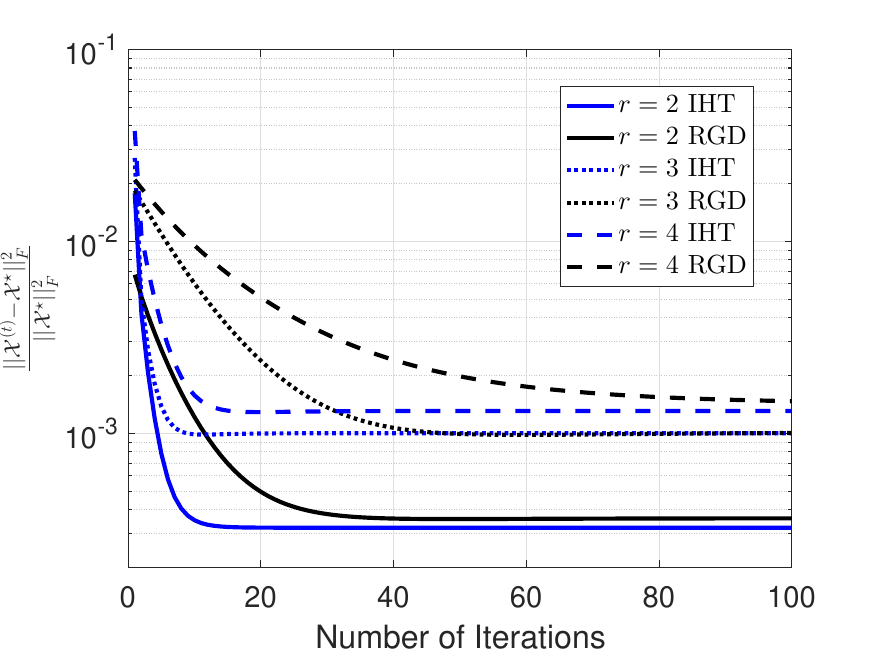}
\end{minipage}
\label{TT_ToT_r}
}
\subfigure[]{
\begin{minipage}[t]{0.3\textwidth}
\centering
\includegraphics[width=5.5cm]{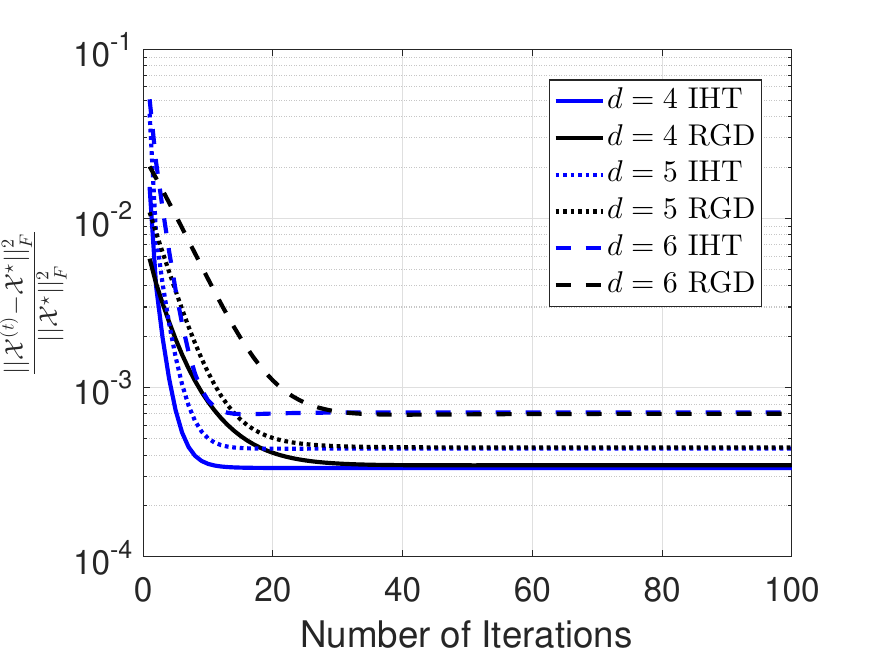}
\end{minipage}
\label{TT_ToT_d}
}
\caption{Convergence for tensor-on-tensor regression (a) for different $N$ with $M=2$, $d = 4$, $r = 2$, $m = 200$ and $\gamma^2 = 0.01$, (b) for different $r$ with $d = 4$, $N = 4$, $M = 2$, $m = 200$ and $\gamma^2 = 0.01$, (c) for different $d$ with $N = 4$, $M = 2$, $r = 2$, $m = 200$ and $\gamma^2 = 0.01$.}
\label{TT ToT convergence}
\end{figure*}

\begin{figure*}[htbp]
\centering
\subfigure[]{
\begin{minipage}[t]{0.3\textwidth}
\centering
\includegraphics[width=5.5cm]{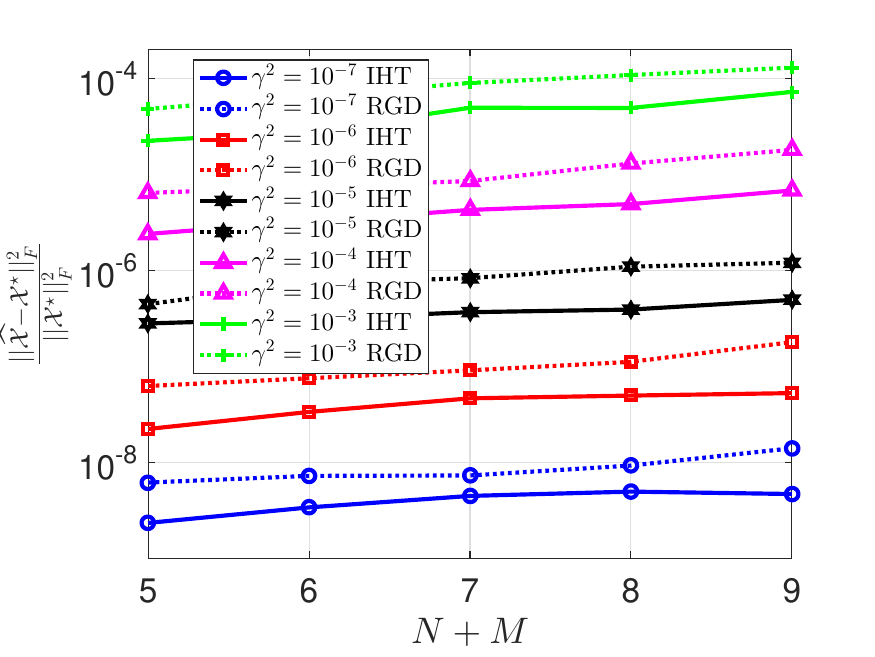}
\end{minipage}
\label{TT_ToT_noise}
}
\subfigure[]{
\begin{minipage}[t]{0.3\textwidth}
\centering
\includegraphics[width=5.5cm]{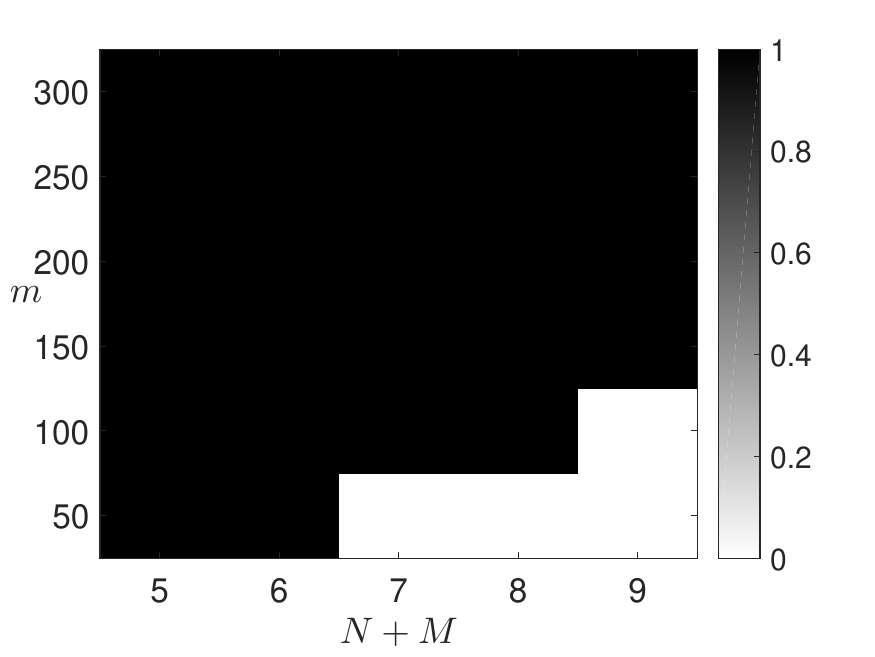}
\end{minipage}
\label{TT ToT_m IHT}
}
\subfigure[]{
\begin{minipage}[t]{0.3\textwidth}
\centering
\includegraphics[width=5.5cm]{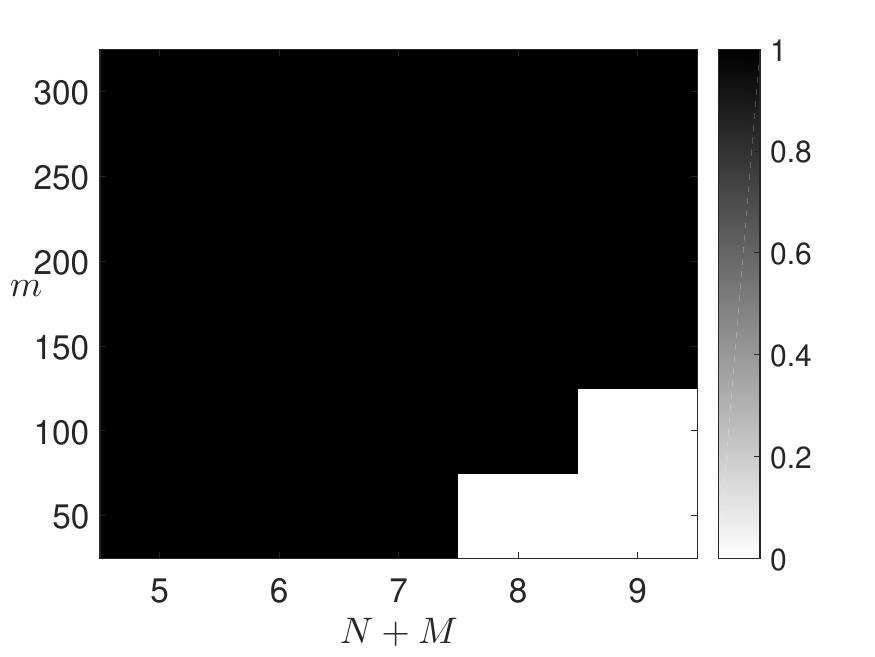}
\end{minipage}
\label{TT_ToT_m RGD}
}
\caption{Performance comparison in the tensor-on-tensor regression, (a) for different $N$ and $\gamma^2$ with $M=2$, $d = 4$, $r = 2$ and $m=200$, (b) (IHT, $\mu = 0.2$) and (c) (RGD, $\mu = 0.5$) for different $N$ and $m$ with $M=2$, $d = 4$, $r = 2$ and $\gamma^2 = 0$.}
\end{figure*}

\subsection{Comparison between IHT and RGD}
\label{Comparison between IHT and RGD}

We now provide further comparisons between IHT and RGD. We begin by analyzing their space and computational complexities, summarized in \Cref{Compelxity comparison among different algorithms}. It is observed that, while IHT incurs a higher space complexity, RGD trades reduced storage requirements for an increased computational burden. Accordingly, the choice between these methods inherently involves a trade-off between memory and computational efficiency, dictated by practical constraints. Moreover, it is worth noting that the terms $m\ol d^N$ and $m(\ol d^N +  \ol d^{N+M} )$ appearing in the complexity expressions stem from the measurement operators $\{ \calB_k \}_{k=1}^m$. In general, such complexities are unavoidable. Nevertheless, when the measurement operators possess a local structure—such as the tensor-product form in quantum state tomography \cite{qin2024quantum,jameson2024optimal}---the exponential complexity can be substantially mitigated to a linear scaling, specifically $O(mN\ol d)$ for space and $O(m(N+M)\ol d \ol r^2)$ for computation.

Next, we proceed to a  comparison of theoretical guarantees.
$(i)$ Regarding initialization, the conditions $\|\calX^{(0)} - \calX^\star\|_F^2$ required for convergence scale as $O(\frac{\underline{\sigma}^2(\calX^\star)}{(N+M)^2})$ for IHT and $O(\frac{\underline{\sigma}^2(\calX^\star)}{(N+M)^4\ol r^2\kappa^2(\calX^\star)})$ for RGD. Notably, the dependency on $N+M$ remains polynomial in both cases, rather than exponential.
$(ii)$ As for convergence rates, IHT achieves a rate of $\frac{2 + \delta_{4\wt r}^2 + 6 \delta_{4\wt r}}{1 + \delta_{4\wt r}^2 + 6 \delta_{4\wt r}}(1 - \mu(1 - \delta_{4\wt r})   )<1$, while RGD converges at a rate of $1 - O(\frac{\underline{\sigma}^2(\calX^\star)}{(N+M)^4\ol r^2\kappa^2(\calX^\star)})$. This indicates that the convergence behavior of IHT is predominantly governed by $\delta_{4\wt r}$, whereas for RGD, the convergence speed is also affected by $N+M$, $\ol r$, and $\kappa(\calX^\star)$, owing to the intrinsic nonconvexity of the optimization landscape. Importantly, both methods maintain only polynomial dependency, ensuring scalability with respect to $\ol d$, $\ol r$, and $N+M$.
$(iii)$ Finally, we consider the asymptotic recovery performance when $t$ becomes large, such that the first term in both \eqref{The expansion of difference two tensors in l2 conclusion theorem} and \eqref{Local Convergence of Riemannian in the noisy ToT regression_Theorem_1 calX} is negligible. In this case, the errors $\|\calX^{(t)} - \calX^\star\|_F^2$ scale as $O\big(\frac{(N+M)\ol d\ol r^2(\log (N+M))\gamma^2 }{m}\big)$ for IHT and $O\big( \frac{(N+M)^5\ol d\ol r^3(\log (N+M)) \kappa^2(\calX^\star)\gamma^2}{m}\big)$ for RGD. The larger error associated with RGD can be attributed to the increased difficulty of navigating the highly nonconvex optimization space, compared to IHT, which can be viewed as solving a convex problem when omitting the projection step.
In conclusion, IHT offers faster convergence, lower computational cost, and improved recovery accuracy at the expense of higher memory usage, whereas RGD provides a more memory-efficient alternative with increased computational burden and larger recovery error. The selection between the two methods should therefore be guided by the specific resource constraints and accuracy requirements of the application.

\subsection{Spectral initialization}
\label{spectral initilazation}
We apply the following spectral initialization to obtain a good initialization for both IHT and RGD
\begin{eqnarray}
    \label{spectral initliazation in ToT regression}
    \calX^{(0)} = \text{SVD}_{\vr}^{tt}\bigg(\frac{1}{m}\calA^*(\calY)\bigg),
\end{eqnarray}
where $\calA^*$ is the adjoint operator of $\calA$ such that the $(s_1,\dots,\\ s_{N+M})$-th element of $\calA^*(\calY)$ can be represented as $\calA^*(\calY)(s_1,\\  \dots, s_{N+M}) = \sum_{k=1}^m \calY(k,s_{N+1},\dots, s_{N+M}) \calB(k,s_1,\dots,s_{N})$. This spectral initialization method has found widespread application in various inverse problems \cite{lu2020phase}, including phase retrieval \cite{candes2015phase,luo2019optimal}, low-rank matrix recovery \cite{Ma21TSP,tong2021accelerating}, and structured tensor recovery \cite{Han20,TongTensor21}. The following result ensures that such an initialization $\calX^{(0)}$ provides a good approximation of $\calX^\star$.

\begin{theorem}
\label{TENSOR SENSING SPECTRAL INITIALIZATION ToT}
Suppose that $\calX^\star$ is in the TT format with ranks $\vr = (r_1,\dots, r_{N + M-1})$.
If $\calA$ satisfies the $3\wt r$-RIP for TT format tensors with a constant $\delta_{3\wt r}$, then with probability at least $1-e^{-\Omega((N+M)\ol d\ol r^2 \log (N+M))}$, the spectral initialization $\calX^{(0)}$ generated by \eqref{spectral initliazation in ToT regression} satisfies
\begin{eqnarray}
    \label{TENSOR SENSING SPECTRAL INITIALIZATION ToT1}
    \hspace{-1cm}\|\calX^{(0)}-\calX^\star\|_F &\!\!\!\!\leq\!\!\!\!& (1+\sqrt{N+M-1}) \bigg(\delta_{3 \wt r}\|\calX^\star\|_F + \nonumber\\
     &\!\!\!\!\!\!\!\!&\hspace{-1.5cm} O\bigg( \frac{\ol r\gamma\sqrt{(1+\delta_{2\wt r})(N+M)\ol d\log (N+M)}}{\sqrt{m}}  \bigg)   \bigg).
\end{eqnarray}
\end{theorem}
The detailed derivation is given in \Cref{proof of spectral init ToT} of the supplementary material.  In words, \Cref{TENSOR SENSING SPECTRAL INITIALIZATION ToT} implies that a sufficiently large $m$ enables the identification of a suitable initialization with the desired proximity to the ground truth.

\section{Experiments}
\label{Sec experiment}

In this section,  we conduct numerical experiments with both synthetic and real data to evaluate the performance of the IHT and RGD algorithms for TT-based tensor-on-tensor regression.

\subsection{Synthetic Data}

We construct an order-$N+M$ ground truth $\calX^\star\in\R^{d_1\times\cdots \times d_{N+M}}$ with ranks $\vr = (r_1,\dots, r_{N+M-1})$ by first generating a random Gaussian tensor with i.i.d. entries from the standard normal distribution, and then using the sequential SVD algorithm to obtain a TT format tensor, which is finally normalized to unit Frobenius norm, i.e., $\|\calX^\star\|_F=1$. Then we generate covariates $\calB$ as a random tensor with i.i.d. entries drawn from the standard normal distribution. Finally, we obtain noisy responses $\calY = \calA(\calX^\star) + \calE$, where the noise $\calE$ is drawn from a Gaussian distribution with a mean of zero and a variance of $\gamma^2$. To simplify the selection of parameters, we set $d=d_1=\cdots=d_{N+M}$ and $r=r_1=\cdots = r_{N+M-1}$. For the IHT and RGD algorithms, we set step size $\mu = 0.5$. For each experimental setting, we conduct $20$ Monte Carlo trials and then take the average over the $20$ trials to report the results.

\paragraph*{Convergence of IHT and RGD} We initially demonstrate the convergence of IHT and RGD in terms of the recovery error $\|\calX^{(t)} - \calX^\star\|_F^2$ across various settings in Figure  \ref{TT ToT convergence}. Notably, we observe a linear convergence rate for both IHT and RGD until reaching a minimum. Furthermore, the plots reveal consistent trends when maintaining a fixed number of training observations $m$ and order $M$ while increasing the values of $N$, $r$, and $d$: $(i)$ the recovery error at initialization using spectral methods increases; $(ii)$ the convergence rate of IHT and RGD becomes slower; $(iii)$ IHT and RGD converge to a solution with larger recovery errors; $(iv)$ the convergence rate of IHT is faster than RGD. These observations are in line with our theoretical findings as presented in \Cref{Convergence analysis of IHT L2}, \Cref{Local Convergence of Riemannian in the noisy ToT regression_Theorem calX} and \Cref{TENSOR SENSING SPECTRAL INITIALIZATION ToT}. In the subsequent experiments, we perform both IHT and RGD for $T = 1000$ iterations
to ensure convergence.

\paragraph*{Stable recovery with noisy responses}  In the second experiment, we will compare the recovery error between IHT and RGD.  We maintain a constant number of training observations $m = 500$ and vary the tensor order $N$ and noise level $\gamma^2$. As illustrated in \Cref{TT_ToT_noise}, the performance of IHT and RGD remains stable as $N$ increases, with the recovery error in the curves exhibiting a polynomial increase. In particular, the recovery error scales approximately linearly with respect to the noise level $\gamma^2$ in line with the second term in \eqref{The expansion of difference two tensors in l2 conclusion theorem} and \eqref{Local Convergence of Riemannian in the noisy ToT regression_Theorem_1 calX}. We also observe that IHT slightly outperforms RGD in terms of recover error, consistent with the comparison between \Cref{Convergence analysis of IHT L2} and \Cref{Local Convergence of Riemannian in the noisy ToT regression_Theorem calX}.

\paragraph*{Exact recovery with clean responses} In the third experiment, we assess the success rate of recovery for IHT and RGD with $\gamma^2 = 0$. The experiments are conducted with varying $N$ and the number of training observations $m$, while keeping $M$ fixed. A successful recovery by IHT/RGD is defined if $\|\wh\calX - \calX^\star \|_F\leq10^{-5}$. We perform $100$ independent trials to evaluate the success rate for each combination of $N$ and $m$. As shown in Figures~\ref{TT ToT_m IHT} and \ref{TT_ToT_m RGD}, the number of training observations $m$ required to ensure exact recovery scales polynomially rather than exponentially with respect to the order $N$, aligning with the findings in our previous analyses.

\begin{table}[!ht]
\begin{center}
\caption{Comparison of IHT and RGD with $m=5000$ in terms of
testing error. 
Here IHT with the full rank refers to the approach without using TT-SVD projection and the RGD with the full rank refers to the method that selects the maximal TT ranks  (which is $(90,219,73)$). The second column represents the TT ranks and the ratio of the memory of the factors to the entire tensor. The best result of each algorithm for each setting (full rank is treated as an individual setting) is in bold.
}
\label{Tu and TT: Accuracy and storage memory comparison in the prediction5000}
\begin{tabular}{|ll|l|lll|}
\hline
\multicolumn{2}{|l|}{\multirow{2}{*}{Methods}} & \multirow{2}{*}{ TT ranks and storage} & \multicolumn{3}{l|}{\hspace{1.8cm}$\gamma^2$}                                                            \\ \cline{4-6}
\multicolumn{2}{|l|}{}                         &                                       & \multicolumn{1}{l|}{0}              & \multicolumn{1}{l|}{1}              & 5              \\ \hline
\multicolumn{2}{|l|}{\multirow{6}{*}{\hspace{0.3cm}IHT}}     & full rank ($100\%$)                   & \multicolumn{1}{l|}{${\bf 0.2957}$}       & \multicolumn{1}{l|}{${\bf 0.3422}$}       & ${\bf 0.523}$
\\ \cline{3-6}
\multicolumn{2}{|l|}{}                         & $(10,10,10)$ ($0.62\%$)               & \multicolumn{1}{l|}{$0.3673$}       & \multicolumn{1}{l|}{$0.3729$}       & $0.398$        \\ \cline{3-6}
\multicolumn{2}{|l|}{}                         & $(20,20,20)$ ($2.28\%$)               & \multicolumn{1}{l|}{$0.3103$}       & \multicolumn{1}{l|}{$0.3247$}       & ${\bf 0.3892}$ \\ \cline{3-6}
\multicolumn{2}{|l|}{}                         & $(30,30,30)$ ($4.99\%$)               & \multicolumn{1}{l|}{$0.2977$}       & \multicolumn{1}{l|}{${\bf 0.3199}$} & $0.4078$       \\ \cline{3-6}
\multicolumn{2}{|l|}{}                         & $(40,40,40)$ ($8.76\%$)               & \multicolumn{1}{l|}{$0.2963$}       & \multicolumn{1}{l|}{$0.3228$}       & $0.4392$       \\ \cline{3-6}
\multicolumn{2}{|l|}{}                         & $(50,50,50)$ ($13.57\%$)              & \multicolumn{1}{l|}{${\bf 0.2956}$} & \multicolumn{1}{l|}{$0.3275$}       & $0.4595$       \\ \hline
\multicolumn{2}{|l|}{\multirow{6}{*}{\hspace{0.3cm}RGD}}     & full rank ($103.46\%$)                   & \multicolumn{1}{l|}{${\bf 0.331}$}        & \multicolumn{1}{l|}{${\bf 0.378}$}        & ${\bf 0.5509}$       \\ \cline{3-6}
\multicolumn{2}{|l|}{}                         & $(10,10,10)$ ($0.62\%$)               & \multicolumn{1}{l|}{$0.4208$}       & \multicolumn{1}{l|}{$0.43$}         & $0.4597$       \\ \cline{3-6}
\multicolumn{2}{|l|}{}                         & $(20,20,20)$ ($2.28\%$)               & \multicolumn{1}{l|}{$0.3764$}       & \multicolumn{1}{l|}{$0.3807$}       & ${\bf 0.4181}$ \\ \cline{3-6}
\multicolumn{2}{|l|}{}                         & $(30,30,30)$ ($4.99\%$)               & \multicolumn{1}{l|}{$0.3722$}       & \multicolumn{1}{l|}{$0.3702$}       & $0.4365$       \\ \cline{3-6}
\multicolumn{2}{|l|}{}                         & $(40,40,40)$ ($8.76\%$)               & \multicolumn{1}{l|}{$0.3484$}       & \multicolumn{1}{l|}{${\bf 0.369}$}  & $0.449$        \\ \cline{3-6}
\multicolumn{2}{|l|}{}                         & $(50,50,50)$ ($13.57\%$)              & \multicolumn{1}{l|}{${\bf 0.3355}$} & \multicolumn{1}{l|}{$0.3713$}       & $0.472$        \\ \hline
\end{tabular}
\end{center}
\end{table}

\subsection{Real Data}
In this experiment, we employ the TT-based ToT model for predicting attributes from facial images, utilizing the Labeled Faces in the Wild database \cite{learned2016labeled}, which has been exploited for ToT regression in \cite{lock2018tensor,llosa2022reduced}. This database comprises over 13,143 publicly available images sourced from the internet, each featuring the face of an individual. Each image is labeled solely with the name of the depicted individual, often a celebrity, and multiple images are available for each person. The attributes encompass characteristics describing the individual (e.g., gender, race, and age), expressions (e.g., smiling, frowning, and eyes open), and accessories (e.g., glasses, make-up, and jewelry). These attributes were determined based on the Faces in the Wild dataset, amounting to a total of 73 attributes measured for each image. Notably, these attributes are measured on a continuous scale; for instance, higher values for the smiling attribute indicate a more pronounced smile, while lower values correspond to no smile.
Specifically, considering that each figure has dimensions $90\times 90\times 3$, the goal is to estimate the unknown tensor $ \calX^\star\in\R^{90\times 90\times3\times73}$. The training and testing covariates are denoted as $\calB_{\text{train}}\in\R^{m\times 90\times 90\times 3}$ and $\calB_{\text{test}}\in\R^{3143\times 90\times 90\times 3}$, respectively.  To further evaluate the denoising capability of the TT structure, a noisy training set $\wt \calY_{\text{train}} = \calY_{\text{train}} + \calE \in\R^{m\times 73}$ is introduced. Here, $\calY_{\text{train}}$ represents the pure facial attributes, and $\calE$ is the noise, with each element following a standard normal distribution $\calN(0,\gamma^2)$. Utilizing the training dataset $\calB_{\text{train}}$ and $\wh \calY_{\text{train}}$, an estimated tensor $\wt \calX$ is obtained. Subsequently, the estimated facial attributes of the testing set are computed as $\wh \calY_\text{test}(s_1,s_2) = \sum_{k_1=1}^{90}\sum_{k_2=1}^{90}\sum_{k_3=1}^{3}\calB_\text{test}(s_1,k_1,k_2,k_3) \wh \calX(k_1,k_2,k_3,s_2)$ using the estimated tensor $\wh\calX$. We compare the performance with the testing error $\frac{\|\wh\calY_{\text{test}} - \calY_{\text{test}}\|_F^2}{\|\calY_{\text{test}}\|_F^2}$.

\begin{table}[!ht]
\begin{center}
\caption{Under the same setting as \Cref{Tu and TT: Accuracy and storage memory comparison in the prediction5000} except for $m=10000$.}
\label{Tu and TT: Accuracy and storage memory comparison in the prediction10000}
\begin{tabular}{|ll|l|lll|}
\hline
\multicolumn{2}{|l|}{\multirow{2}{*}{Methods}} & \multirow{2}{*}{ TT ranks and storage} & \multicolumn{3}{l|}{\hspace{1.8cm}$\gamma^2$}                                                            \\ \cline{4-6}
\multicolumn{2}{|l|}{}                         &                                       & \multicolumn{1}{l|}{0}              & \multicolumn{1}{l|}{1}              & 5              \\ \hline
\multicolumn{2}{|l|}{\multirow{6}{*}{\hspace{0.3cm}IHT}}     & full rank ($100\%$)                   & \multicolumn{1}{l|}{${\bf 0.2815}$}       & \multicolumn{1}{l|}{${\bf 0.3056}$}       & ${\bf 0.3987}$
\\ \cline{3-6}
\multicolumn{2}{|l|}{}                         & $(10,10,10)$ ($0.62\%$)               & \multicolumn{1}{l|}{$0.3632$}       & \multicolumn{1}{l|}{$0.3675$}       & $0.3767$        \\ \cline{3-6}
\multicolumn{2}{|l|}{}                         & $(20,20,20)$ ($2.28\%$)               & \multicolumn{1}{l|}{$0.299$}       & \multicolumn{1}{l|}{$0.3054$}       & ${\bf 0.3382}$ \\ \cline{3-6}
\multicolumn{2}{|l|}{}                         & $(30,30,30)$ ($4.99\%$)               & \multicolumn{1}{l|}{$0.2859$}       & \multicolumn{1}{l|}{${\bf 0.2973}$} & $0.3422$       \\ \cline{3-6}
\multicolumn{2}{|l|}{}                         & $(40,40,40)$ ($8.76\%$)               & \multicolumn{1}{l|}{$0.2832$}       & \multicolumn{1}{l|}{$0.2979$}       & $0.356$       \\ \cline{3-6}
\multicolumn{2}{|l|}{}                         & $(50,50,50)$ ($13.57\%$)              & \multicolumn{1}{l|}{${\bf 0.2821}$} & \multicolumn{1}{l|}{$0.2997$}       & $0.3676$       \\ \hline
\multicolumn{2}{|l|}{\multirow{6}{*}{\hspace{0.3cm}RGD}}     & full rank ($103.46\%$)                   & \multicolumn{1}{l|}{${\bf 0.3109}$}        & \multicolumn{1}{l|}{${\bf 0.3455}$}        & ${\bf 0.4341}$       \\ \cline{3-6}
\multicolumn{2}{|l|}{}                         & $(10,10,10)$ ($0.62\%$)               & \multicolumn{1}{l|}{$0.419$}       & \multicolumn{1}{l|}{$0.4274$}         & $0.4388$       \\ \cline{3-6}
\multicolumn{2}{|l|}{}                         & $(20,20,20)$ ($2.28\%$)               & \multicolumn{1}{l|}{$0.3621$}       & \multicolumn{1}{l|}{$0.3617$}       & $0.3928$
\\ \cline{3-6}
\multicolumn{2}{|l|}{}                         & $(30,30,30)$ ($4.99\%$)               & \multicolumn{1}{l|}{$0.3399$}       & \multicolumn{1}{l|}{$0.3511$}       & ${\bf 0.381}$       \\ \cline{3-6}
\multicolumn{2}{|l|}{}                         & $(40,40,40)$ ($8.76\%$)               & \multicolumn{1}{l|}{$0.3278$}       & \multicolumn{1}{l|}{$0.3427$}  & $0.3887$
\\ \cline{3-6}
\multicolumn{2}{|l|}{}                         & $(50,50,50)$ ($13.57\%$)              & \multicolumn{1}{l|}{${\bf 0.315}$} & \multicolumn{1}{l|}{${\bf 0.336}$}       & $0.3974$        \\ \hline
\end{tabular}
\end{center}
\end{table}

Next, we will assess the performance of IHT and RGD with different ranks. As a baseline comparison, we also include results without low-rank structure, utilizing IHT and RGD, respectively. For IHT, we use a step size of $\mu = 5\times10^{-4}$ and perform 1500 iterations. In the case of RGD, a step size of $\mu = 4\times10^{-4}$ is applied with a total of 3000 iterations. Since the ground-truth tensor $\calX^\star$ is unknown and the computational complexity of $\ol \sigma^2(\calX^{(t)})$ is high, we use $\|\calX^{(t)}\|_F^2$ as the balancing parameter in the RGD algorithm. In Tables~\ref{Tu and TT: Accuracy and storage memory comparison in the prediction5000} and \ref{Tu and TT: Accuracy and storage memory comparison in the prediction10000}, we present the results for IHT, and RGD with $m=5000$ and $m=10000$. The running time (s) per iteration for IHT and RGD with the full rank, IHT (TT ranks: $10-50$) and GD (TT ranks: $10-50$) are $0.79$, $0.99$, $[0.85 \ 0.99]$ and $[0.85 \ 0.88]$ for $m = 5000$, and $1.68$, $1.86$, $[1.72 \  1.87]$ and $[1.65 \ 1.76]$ respectively for $m = 10000$. Here $[a \ b]$ is the time range.  According to these two tables: $(i)$ The recovery accuracy of IHT is superior to RGD for various values of $\gamma^2$ and $\vr$. This is further substantiated by the results in \Cref{Convergence analysis of IHT L2} and \Cref{Local Convergence of Riemannian in the noisy ToT regression_Theorem calX}, where the recovery error associated with IHT consistently outperforms that of RGD. $(ii)$ The number of iterations required by RGD surpasses that of IHT, attributable to its slower convergence rate. This conclusion is reinforced by the findings in \Cref{Convergence analysis of IHT L2} and \Cref{Local Convergence of Riemannian in the noisy ToT regression_Theorem calX}, where the convergence rate of IHT is shown to be superior to that of RGD. $(iii)$ In cases where the training data is noiseless ($\gamma^2 = 0$), the performance of IHT and RGD with $\vr = (50,50,50)$ approaches that of IHT and RGD without low-rank structure, respectively. It is noteworthy that the storage memory occupies only $13.57\%$ of the entire tensor. $(iv)$ For $\gamma^2 =1$ or $5$, the TT structure demonstrates better denoising capabilities compared to the entire tensor. Additionally, with increasing values of $\gamma^2$ and $\vr$, the recovery accuracy tends to decrease, but an optimal performance is achieved with a small $\vr$. $(v)$ With an increase in the number of training observations $m$, both recovery error and denoising improve. This observation is consistent with the results presented in \Cref{Comparison among different tensor regression}, further corroborating the robustness of the TT structure under varying conditions.

\section{Conclusion}
\label{conclusion}

In this paper, we study the theoretical and algorithmic aspects of the TT-based ToT regression model. We establish statistically optimal error bounds, encompassing both upper and minimax lower bounds, for incorporating TT structure in the general ToT regression problem. We then exploit two complementary optimization algorithms to solve these problems: the Iterative Hard Thresholding (IHT) that optimizes the entire tensor and the Riemannian Gradient Descent (RGD) that optimizes over the factors. When the RIP is satisfied, spectral initialization aids in obtaining a suitable starting point, and we further prove the linear convergence rate of both the IHT and RGD. Additionally, we demonstrate that the IHT can achieve the same order as the theoretical upper error bound, whereas the RGD, although trailing the IHT, can still attain a polynomial error bound relative to the tensor order. Experimental results validate our theoretical and algorithmic analyses. This study not only enhances the applicability and reliability of the TT-based ToT regression but also paves the way for further exploration in tensor decompositions and regression models.

As previously discussed, while RGD offers advantages for high-order tensor problems due to its lower storage complexity, it suffers from a slower convergence rate and a larger recovery error compared to IHT, largely attributed to the condition number $\kappa(\calX^\star)$; in particular, a larger $\kappa(\calX^\star)$ exacerbates both convergence speed and recovery performance.
To address this issue, a promising future direction is to incorporate scaling techniques \cite{tong2021accelerating, tong2021low, TongTensor21} into the proposed algorithms, aiming to mitigate the detrimental effects of $\kappa(\calX^\star)$ on convergence speed and recovery accuracy.
Furthermore, although the current theoretical guarantees are established under subgaussian measurement assumptions, it remains an open question whether similar conclusions hold under more practical scenarios, such as quantum measurements \cite{qin2024quantum, jameson2024optimal, qin2024sample, qin2024optimal}, where the RIP condition typically does not apply.

Another important direction is to generalize the proposed framework to the Hierarchical Tucker format \cite{grasedyck2010hierarchical, da2015optimization} and more general tensor network structures \cite{cichocki2016tensor, orus2019tensor} within the ToT regression model. In contrast to TT decomposition, the Hierarchical Tucker format lacks a fixed compression structure, while general tensor networks often involve loops, posing substantial algorithmic challenges and necessitating the development of new methodologies. Furthermore, exploring the extension of TT-based ToT regression to neural network models \cite{kossaifi2020tensor} represents another promising avenue for future research.




\onecolumn

\clearpage
\begin{center}
    \LARGE \textbf{Supplementary Material}
\end{center}

\section{Tensor Contraction for TT-ToT Model}
\label{Intro TT ToT model}

In this section, we will define tensor contraction and utilize it to establish the concise representations of TT and ToT regression models. Firstly, tensor contraction is defined as $\calA\times_{i}^{j}\calB\in\R^{d_1\times\cdots \times d_{i-1}  \times d_{i+1}\times \cdots\times d_{N} \times h_1 \times \cdots \times h_{j-1} \times h_{j+1}\times \cdots\times h_{M}}$ with $(s_1,\dots,s_{i-1},s_{i+1},\dots,s_N,f_1,\dots,f_{j-1},f_{j+1},\dots,f_M)$-th element being $ \sum_{k}\calA(s_1,\dots,s_{i-1},k,s_{i+1},\dots,s_N)\calB(f_1, \dots, \\ f_{j-1},k,f_{j+1},\dots,f_M)$ for $\calA\in\R^{d_1 \times \cdots \times d_{i-1}\times d_k \times d_{i+1}\times \cdots\times d_{N}}$ and $\calB\in\R^{h_1 \times \cdots \times h_{j-1}\times d_k \times h_{j+1}\times \cdots\times h_{M}}$.
The procedure of $\calA\times_{i_1,\dots,i_n}^{j_1,\dots,j_n}\calB$ can be viewed as a sequence of $n$ tensor contractions $\calA\times_{i_k}^{j_k}\calB, k\in[n]$.

Based on the tensor contraction, we can simplify the TT format as
\begin{eqnarray}
    \label{Definition of Tensor Train1 appe}
    \calX =[\mX_1,\dots, \mX_{N+M}]  = \mX_1\times_{2}^1\mX_2\times_{3}^1\cdots \times_{N+M}^1 \mX_{N+M}.\nonumber
\end{eqnarray}
Here the tensor contraction $\mX_1\times_2^1\mX_2$ results in a new tensor $\calX_1$ of size $d_1\times d_2\times r_2$ with $(s_1,s_2,f_2)$-th element being $\sum_{f_1}\mX_1(s_1,f_1)\mX_2(f_1,s_2,f_2)$. Likewise, $\calX_1\times_3^1\mX_3$ results in a tensor of size $d_1\times d_2 \times d_3 \times r_3$ with $(s_1,s_2, s_3, f_3)$-th element being $\sum_{f_2}\calX_1(s_1,s_2,f_2)\mX_3(f_2,s_3,f_3)$.

In addition, the ToT regression model can be written succinctly as
\begin{eqnarray}
    \label{The compact form of ToT model appe}
    \calY = \calA(\calX^\star) + \calE = \calB\times_{2,\dots,N+1}^{1,\dots,N}\calX^\star + \calE ,
\end{eqnarray}
where the $(k,s_{N+1},\dots,s_{N+M})$-th element in $\calB\times_{2,\dots,N+1}^{1,\dots,N}\calX^\star$ is computed via $\sum_{s_1,\dots,s_N} \calB(k,s_1,\dots, s_N)  \calX^\star(s_1,\dots, s_N, \\  s_{N+1}, \dots,s_{N+M})$.

In the following subsections, we will employ tensor contraction to simplify presentation.

\section{Proof of \Cref{RIP condition fro the ToT TT regression Lemma}}
\label{Proof of ToT RIP}

\begin{proof}
According to \cite{qin2024quantum,Rauhut17}, when \eqref{eq:mrip ToT TT} is satisfied, with probability $1-\epsilon$, we have
\begin{eqnarray}
    \label{RIP condition fro the TT regression l2}
     (1-\delta_{\wt r})\|\calX(:,\dots,:,s_{N+1},\dots,s_{N+M})\|_F^2&\!\!\!\!\leq\!\!\!\!& \frac{1}{m}\|\mathcal{A}(\calX)(:,s_{N+1},\dots,s_{N+M})\|_2^2\nonumber\\
    &\!\!\!\!\!\!\!\!&\leq(1+\delta_{\wt r})\|\calX(:,\dots,:,s_{N+1},\dots,s_{N+M})\|_F^2,\nonumber\\
    &\!\!\!\!\!\!\!\!& s_{j}\in[d_j], j=N+1,\dots, N+M,
\end{eqnarray}
which together with the facts that $\|\mathcal{A}(\calX)\|_F^2 = \sum_{s_{N+1},\dots, s_{N+M}}\|\mathcal{A}(\calX)(:,s_{N+1},  \dots,s_{N+M})\|_2^2$ and $\|\calX\|_F^2 =  \sum_{s_{N+1},\dots, s_{N+M}} \|\calX(:,\dots,:,s_{N+1},  \dots,s_{N+M})\|_F^2$ completes the proof.
\end{proof}

\section{Proof of \Cref{Upper bound of error difference}}
\label{Proof of upper bound for error difference}

According to \eqref{wX and X star relationship_1}, the proof is mainly to bound $\frac{2}{m}\< \calE, \calA(\wh\calX - \calX^\star) \>$. We begin by expressing it as follows:
\begin{eqnarray}
    \label{upper bound of difference}
    \frac{2}{m}\<  \calE, \calA(\wh\calX - \calX^\star) \> &\!\!\!\!\!\!=\!\!\!\!\!\!& \frac{2\|\wh\calX - \calX^\star\|_F}{m}\max_{\calH\in\setX_{2\ol{r}}, \|\calH\|_F\leq 1} \<\calH, \calA^*(\calE) \>\nonumber\\
    &\!\!\!\!\!\!=\!\!\!\!\!\!& \frac{2\|\wh\calX - \calX^\star\|_F}{m}\hspace{-0.5cm}  \max_{[\mH_1,\dots,\mH_{N+M}]\in\setX_{2\ol{r}},\atop \|\mH_i\|_F\leq 1,i\in[N+M]} \hspace{-0.2cm} \<[\mH_1,\dots,\mH_{N+M}], \calA^*(\calE) \>,
\end{eqnarray}
where we define $\calA^*$ as the adjoint operator of $\calA$ and each element in $\calA^*(\calE)$ is computed via $\calA^*(\calE)(s_1,\dots,s_{N+M}) = \sum_{k=1}^m \calE(k,s_{N+1},\dots, s_{N+M}) \calB(k,s_1,\dots,s_{N})$.

Next, according to \cite{zhang2018tensor}, for each $i \in [N + M]$, we can construct an $\epsilon$-net $\{L(\mH_i^{(1)}), \dots, L(\mH_i^{(n_i)})  \}$ with the covering number $n_i\leq (\frac{2+\epsilon}{\epsilon})^{d_ir_{i-1}r_i}$ for the set of factors $\{L(\mH_i)\in\R^{d_ir_{i-1}\times r_i}: \|L(\mH_i)\|_2\leq 1\}$ such that
\begin{eqnarray}
    \label{ProofOf<H,X>forSubGaussian_proof1}
    \sup_{L(\mH_i): \|L(\mH_i)\|_2\leq 1}\min_{p_i\leq n_i} \|L(\mH_i)-L(\mH_i^{(p_i)})\|_2\leq \epsilon.
\end{eqnarray}
Therefore, we can construct an $\epsilon$-net $\{\calH^{(1)},\! \ldots, \! \calH^{(n_1\cdots n_{N+M}) }  \}$ with covering number
\begin{eqnarray}
    \label{Total number of covering number}
    \Pi_{i=1}^{N+M} n_i \leq \bigg(\frac{2+\epsilon}{\epsilon}\bigg)^{\sum_{i=1}^{N+M}d_ir_{i-1}r_i}\!\!\!\!\!\! \leq \bigg(\frac{2+\epsilon}{\epsilon}\bigg)^{(N+M)\ol d\ol r^2 }
\end{eqnarray}
for any TT format tensors $\calH = [\mH_1,\dots, \mH_{N+ M}]\in\R^{d_1\times \cdots \times d_{N+ M}}$ with TT ranks $(r_1,\dots, r_{N+ M-1})$ and $r_0 = r_{N+M} = 1$. Note that $\ol r=\max_{i=1}^{N + M-1}r_i$ and $\ol d=\max_{i=1}^{N+ M}d_i$.

Denote by $T$ the value of \eqref{upper bound of difference}, i.e.,
\begin{eqnarray}
    \label{ProofOf<H,X>forSubGaussian_proof3}
    [\widetilde\mH_1,\dots, \widetilde\mH_{N+M}] &\!\!\!\!=\!\!\!\!& \hspace{-0.2cm} \argmax_{{[\mH_1,\dots,\mH_{N+M}]\in\setX_{2\ol{r}},\atop \|\mH_i\|_F\leq 1,i\in[N+M]}}\frac{1}{m} \<[\mH_1,\dots, \mH_{N+M}], \calA^*(\calE) \>,\\
    \label{ProofOf<H,X>forSubGaussian_proof4}
    T:&\!\!\!\! = \!\!\!\!& \frac{1}{m}\<\calA^*(\calE), [\widetilde\mH_1,\dots, \widetilde\mH_{N+M}]  \>,
\end{eqnarray}
where $\calA^*(\calE) = \calB\times_1^1 \calE$. Using $\calI$ to denote the index set $[n_1]\times \cdots \times [n_{N+M}]$, then according to the construction of the $\epsilon$-net, there exists $p=(p_1,\dots, p_{N+M})\in\calI$ such that
\begin{eqnarray}
    \label{ProofOf<H,X>forSubGaussian_proof5}
    \|\widetilde\mH_i - \mH_i^{(p_i)} \|_F\leq\epsilon, \ \  i\in [N+M],
\end{eqnarray}
and taking $\epsilon=\frac{1}{2(N+M)}$ gives
\begin{eqnarray}
    \label{ProofOf<H,X>forSubGaussian_proof6}
    T&\!\!\!\!=\!\!\!\!&\frac{1}{m}\<\calA^*(\calE), [\mH_1^{(p_1)},\dots, \mH_{N+M}^{(p_{N+M})}]  \>+ \frac{1}{m}\<\calA^*(\calE),   [\widetilde\mH_1,\dots, \widetilde\mH_{N+M}] - [\mH_1^{(p_1)},\dots, \mH_{N+M}^{(p_{N+M})}]  \>\nonumber\\
    &\!\!\!\!=\!\!\!\!&\frac{1}{m}\<\calA^*(\calE), [\mH_1^{(p_1)},\dots, \mH_{N+M}^{(p_{N+M})}]  \>+ \frac{1}{m}\<\calA^*(\calE),   \sum_{a_1=1}^{N+M}[\mH_1^{(p_1)},\dots, \mH_{a_1}^{(p_{a_1})}-\widetilde\mH_{a_1},  \dots, \widetilde\mH_{N+M}]\>\nonumber\\
    &\!\!\!\!\leq\!\!\!\!&\frac{1}{m}\<\calA^*(\calE), [\mH_1^{(p_1)},\dots, \mH_{N+M}^{(p_{N+M})}]  \> + (N+M) \epsilon T\nonumber\\
    &\!\!\!\!=\!\!\!\!&\frac{1}{m}\<\calA^*(\calE), [\mH_1^{(p_1)},\dots, \mH_{N+M}^{(p_{N+M})}]  \>+\frac{T}{2},
\end{eqnarray}
where the second line uses \Cref{EXPANSION_A1TOAN-B1TOBN_1} to rewrite $[\widetilde\mH_1,\dots, \widetilde\mH_{N+M}] - [\mH_1^{(p_1)},\dots, \mH_{N+M}^{(p_{N+M})}]$ into a sum of $N+ M$ terms. The first inequality follows $\|[\mH_1^{(p_1)},\dots, \mH_{a_1}^{(p_{a_1})}-\widetilde\mH_{a_1},  \dots, \widetilde\mH_{N+M}]\|_F\leq \|\mH_1^{(p_1)}\|_F\cdots  \|\mH_{a_1}^{(p_{a_1})}- \widetilde\mH_{a_1}\|_F \cdots \|\widetilde\mH_{N+M}\|_F$.

Note that each element in  $\calE$ follows the normal distribution $\calN(0,\gamma^2)$. When conditional on $\calB$, for any fixed $\calH^{(p)}\in\R^{d_1\times\cdots\times d_{N+M}}$, $\frac{1}{m}\<\calA^*(\calE),  \calH^{(p)} \>$ has normal distribution with zero mean and variance $\frac{\gamma^2\|\calA(\calH^{(p)})\|_F^2}{m^2}$, which implies that
\begin{eqnarray}
    \label{the tail function of fixed gaussian random variable}
    \P{\frac{1}{m}|\<\calA^*(\calE),  \calH^{(p)} \>| \geq t | \calB}\leq e^{-\frac{m^2t^2}{2\gamma^2\|\calA(\calH^{(p)})\|_F^2}}.
\end{eqnarray}
Furthermore, under the event $F:=\{\calA$   satisfies $2\wt r$-RIP with  constant  $\delta_{2\wt r}$ $\}$, it holds that  $\frac{1}{m}\|\calA(\calH^{(p)})\|_F^2\leq(1+\delta_{2\wt r})\|\calH^{(p)}\|_F^2$. Plugging this together with the fact $\|\calH^{(p)}\|_F\leq 1$ obtained from  $\|\mH_i^{(p_i)}\|_F\leq 1, i \in [N+M]$ into the above further gives
\begin{eqnarray}
    \label{the tail function of fixed gaussian random variable1}
    \P{\frac{1}{m}|\<\calA^*(\calE),  \calH^{(p)} \>| \geq t | F}\leq e^{-\frac{mt^2}{2(1+\delta_{2\wt r})\gamma^2}}.
\end{eqnarray}

We now apply this tail bound to \eqref{ProofOf<H,X>forSubGaussian_proof6} and get
\begin{eqnarray}
    \label{the tail function of fixed gaussian random variable 2}
\P{T \geq t | F} &\!\!\!\!\leq \!\!\!\!& \P{ \max_{p_1,\dots, p_{N+M}} \frac{1}{m}\<\calA^*(\calE), [\mH_1^{(p_1)},\dots, \mH_{N+M}^{(p_{N+M})}]  \> \geq \frac{t}{2} | F}\nonumber\\
    &\!\!\!\!\leq\!\!\!\!& \bigg(\frac{2+\epsilon}{\epsilon}\bigg)^{4(N+M)\ol d\ol r^2 }e^{-\frac{mt^2}{8(1+\delta_{2\wt r})\gamma^2}}\nonumber\\
    &\!\!\!\!\leq\!\!\!\!& e^{-\frac{mt^2}{8(1+\delta_{2\wt r})\gamma^2} + c_1(N+M)\ol d\ol r^2 \log (N+M)},
\end{eqnarray}
where $c_1$ is a constant and based on the assumption in \eqref{ProofOf<H,X>forSubGaussian_proof6} that $\frac{2+\epsilon}{\epsilon}=\frac{2+\frac{1}{2(N+M)}}{\frac{1}{2(N+M)}}=4(N+M)+1$.

Hence, we can take $t = \frac{c_2\ol r\sqrt{(1+\delta_{2\wt r})(N+M)\ol d(\log (N+M))}}{\sqrt{m}}\gamma$ with a constant $c_2$ and further derive
\begin{eqnarray}
    \label{the tail function of fixed gaussian random variable 3}
    &\!\!\!\!\!\!\!\!&\P{T \leq \frac{c_2\ol r\sqrt{(1+\delta_{2\wt r})(N+M)\ol d(\log (N+M))}}{\sqrt{m}}\gamma } \nonumber\\
    &\!\!\!\!\geq\!\!\!\!& \P{T \leq \frac{c_2\ol r\sqrt{(1+\delta_{2\wt r})(N+M)\ol d(\log (N+M))}}{\sqrt{m}}\gamma \cap F } \nonumber\\
    &\!\!\!\!\geq\!\!\!\!& P(F) \P{T \leq \frac{c_2\ol r\sqrt{(1+\delta_{2\wt r})(N+M)\ol d(\log (N+M))}}{\sqrt{m}}\gamma |F }\nonumber\\
    &\!\!\!\!\geq\!\!\!\!& (1-e^{-c_3N\wt d\wt r^2 \log N})(1-e^{-c_4(N+M)\ol d\ol r^2 \log (N+M)})\nonumber\\
    &\!\!\!\!\geq\!\!\!\!&  1-e^{-c_3N\wt d\wt r^2 \log N} -e^{-c_4(N+M)\ol d\ol r^2 \log (N+M)},
\end{eqnarray}
where $c_i,i=3,4$ are constants. Note that $P(F)$ is obtained via \Cref{RIP condition fro the ToT TT regression Lemma} by setting $\epsilon$ in \eqref{eq:mrip ToT TT} to be $e^{-c_3N\wt d\wt r^2 \log N}$.

Combing $\frac{1}{m}\|\calA(\wh\calX  -  \calX^\star)\|_F^2 \geq (1-\delta_{2\wt r})\|\wh\calX  -  \calX^\star\|_F^2$, we can obtain
\begin{eqnarray}
    \label{upper bound of error final}
    \|\wh \calX - \calX^\star\|_F\leq O\bigg(\frac{\ol r\gamma\sqrt{(1+\delta_{2\wt r})(N+M)\ol d(\log (N+M))}}{(1-\delta_{2\wt r})\sqrt{m}}\bigg).\nonumber
\end{eqnarray}

\section{Proof of \Cref{minimax bound of error difference}}
\label{proof of minimax bound for TT}

Suppose we can find a set of $\{\calX^i  \}_{i=1}^n\in\setX_{\ol{r}}$ such that $\min_{i\neq j}\|\calX^i - \calX^j\|_F\geq s$.
According to \cite[Theorem 4]{luo2022tensor}, when each element of $\calB$ and $\calE$ respectively follow $\calN(0,1)$ and $\calN(0,\gamma^2)$, we have
\begin{eqnarray}
    \label{lower bound of the minimax}
    \inf_{\wh \calX}\sup_{\calX\in\setX_{\ol{r}}}\E{\|\wh{\calX} - \calX\|_F} \geq \frac{s}{2}\bigg(1- \frac{\frac{m}{2\gamma^2}\max_{i_1\neq i_2}\|\calX^{i_1} -\calX^{i_2} \|_F^2 + \log 2 }{\log n}    \bigg).
\end{eqnarray}

Next, we consider one construction for the sets of $\{\calX^i  \}_{i=1}^n\in\setX_{\ol{r}}$ such that we can obtain a proper lower bound for $\min_{i_1\neq i_2}\|\calX^{i_1} -\calX^{i_2} \|_F$ and a proper upper bound for $\max_{i_1\neq i_2}\|\calX^{i_1} -\calX^{i_2} \|_F$.
Because of $\calX^{i_1}, \calX^{i_2}\in\setX_{\ol r}$, we can utilize the TT-SVD \cite{Oseledets11,holtz2012manifolds} to construct orthonormal formats $\calX^{i_1} = [\wt\mX_1,\dots, \wt\mX_{j-1}, \wt\mX_j^{i_1}, \wt\mX_{j+1}, \dots, \wt\mX_{N+M} ]$ and $ \calX^{i_2} = [\wt\mX_1,\dots, \wt\mX_{j-1}, \wt\mX_j^{i_2}, \wt\mX_{j+1}, \dots, \wt\mX_{N+M} ]$ with orthonormal matrices $L(\wt\mX_k), k< j $ and $R(\wt\mX_k), k>j $, where $L(\wt\mX_k)$ is defined in \eqref{left orthogonal form L} and  $R(\wt\mX_k)=\begin{bmatrix}\wt\mX_k(1) &  \cdots &  \wt\mX_k(d_k) \end{bmatrix}$. Given any $\delta >0 $ and $\min r_i\geq C'$ by \cite[Lemma 5]{agarwal2012noisy}, we can further construct a set of  $\{ \wt\mX_j^{1}, \dots, \wt\mX_j^{n_j} \}$  with cardinality $n_j\geq \frac{1}{4}e^{\frac{d_jr_{j-1}r_j}{128}}$ such that: (1) $\| \wt\mX_j^{i}\|_F = \delta$ holds for all $i = 1,\dots, n_j$, (2) $\| \wt\mX_j^{i_1} - \wt\mX_j^{i_2}\|_F \geq \delta$ for all $i_1,i_2\in[n_j], i_1\neq i_2$.

Since we have $\|\calX^{i_1} - \calX^{i_2}\|_F = \| [ \wt\mX_1, \dots,\wt\mX_j^{i_1} - \wt\mX_j^{i_2}, \dots,\wt\mX_{N+M} ] \|_F  = \|\wt\mX_j^{i_1} - \wt\mX_j^{i_2}\|_F$, we can get
\begin{eqnarray}
    \label{two bounds in the minimax}
    &&\max_{i_1\neq i_2}\|\calX^{i_1} - \calX^{i_2}\|_F = \max_{i_1\neq i_2}\|\wt\mX_j^{i_1} - \wt\mX_j^{i_2}\|_F\leq 2\delta,\nonumber\\
    &&\min_{i_1\neq i_2}\|\calX^{i_1} - \calX^{i_2}\|_F = \min_{i_1\neq i_2}\|\wt\mX_j^{i_1} - \wt\mX_j^{i_2}\|_F\geq \delta.
\end{eqnarray}
Then we plug \eqref{two bounds in the minimax} into \eqref{lower bound of the minimax} and have
\begin{eqnarray}
    \label{lower bound of the minimax conlcusion}
    \inf_{\wh \calX}\sup_{\calX\in\setX_{\ol{r}}}\E{\|\wh{\calX} - \calX\|_F} &\!\!\!\!\geq\!\!\!\!& \frac{\delta}{2}\bigg(1- \frac{\frac{2m\delta^2}{\gamma^2} + \log 2 }{c_1d_jr_{j-1}r_j}    \bigg)\nonumber\\
    &\!\!\!\!\geq\!\!\!\!& c_2 \sqrt{\frac{d_jr_{j-1}r_j}{m}}\gamma,
\end{eqnarray}
where $c_1, c_2$ are positive constants and the second inequality follows by substituting $\delta = c_3\sqrt{\frac{d_jr_{j-1}r_j}{m}}\gamma$ with a constant $c_3>0$ into the first inequality.

Finally, following the procedure in \cite[Theorem 4]{luo2022tensor}, we sum all cases for $j\in[N+M]$ and calculate the average to derive
\begin{eqnarray}
    \label{lower bound of the minimax conlcusion ave}
    \inf_{\wh \calX}\sup_{\calX\in\setX_{\ol{r}}}\E{\|\wh{\calX} - \calX\|_F} \geq C\sqrt{\frac{\sum_{j=1}^{N+M}d_jr_{j-1}r_j}{m}}\gamma,
\end{eqnarray}
where $C$ is a universal constant.

\section{Proof of \Cref{Convergence analysis of IHT L2}}
\label{proof of converegence L2 IHT}

\begin{proof}
Before proving the convergence rate of the IHT, we need to define one restricted Frobenius norm. Inspired by \cite{zhang2021preconditioned} and \cite[Lemma 20]{tong2022scaled}, for any tensor $\calH\in\R^{d_1\times\cdots\times d_{N+M}}$, the definition of restricted Frobenius norm is
\begin{eqnarray}
    \label{Definition of the restricted F norm}
    \|\calH\|_{F,\ol r} &\!\!\!\!=\!\!\!\!&\max_{i\in[N+M-1]} \sqrt{\sum_{j=1}^{r_i}\sigma_j^2(\calH^{\<i \>})} \nonumber\\
    &\!\!\!\!=\!\!\!\!& \max_{\mV_i\in\R^{d_{i+1}\cdots d_{N+M}\times r_i}, \atop \mV_i\mV_i^\top = \mId_{r_i}, i\in[N+M-1]}\|\calH^{\< i\>} \mV_i \|_F   \nonumber\\
    &\!\!\!\!=\!\!\!\!& \max_{\calX\in\setX_{\ol{r}}, \|\calX\|_F\leq 1} |\<\calH,  \calX \>|.
\end{eqnarray}

By \Cref{Perturbation bound for TT SVD}, we expand $\|\calX^{(t+1)} - \calX^\star\|_F^2$ as following:
\begin{eqnarray}
    \label{The MSE relationship of different time in l2}
    &\!\!\!\!\!\!\!\!&\|\calX^{(t+1)} - \calX^\star\|_F^2 = \|\calX^{(t+1)} - \calX^\star\|_{F,2\ol r}^2\nonumber\\
     &\!\!\!\!\leq\!\!\!\!& \bigg(1+ \frac{600(N+M)}{\underline{\sigma}({\calX^\star})}\|\calX^{(0)} - \calX^\star \|_F \bigg)\| {\widetilde{\calX}}^{(t)}   - \calX^\star \|_{F,2\ol r}^2\nonumber\\
    &\!\!\!\!=\!\!\!\!&\bigg(1+ \frac{600(N+M)}{\underline{\sigma}({\calX^\star})}\|\calX^{(0)} - \calX^\star \|_F \bigg) \big(\|\calX^{(t)} - \calX^\star\|_F^2  -2\mu\< \nabla g(\calX^{(t)}), \calX^{(t)} - \calX^\star \> + \mu^2\|\nabla g(\calX^{(t)})\|_{F,2\ol r}^2\big),
\end{eqnarray}
where ${\widetilde{\calX}}^{(t)} = \calX^{(t)} - \mu\nabla g(\calX^{(t)})$ and the gradient is defined as follows:
\begin{eqnarray}
    \label{The definition of gradient in the IHT TOT}
    \nabla g(\calX^{(t)}) &\!\!\!\!=\!\!\!\!& \frac{1}{m}\calA^*(\calA(\calX^{(t)}) - \calY)\nonumber\\
    &\!\!\!\!=\!\!\!\!& \frac{1}{m} \calB\times_1^1 (\calA(\calX^{(t)}) - \calY).
\end{eqnarray}

First, based on the RIP, we have
\begin{eqnarray}
    \label{lower bound of cross term in IHT l2}
    &\!\!\!\!\!\!\!\!&\< \nabla g(\calX^{(t)}), \calX^{(t)} - \calX^\star \>\nonumber\\
    &\!\!\!\! = \!\!\!\!& \frac{1}{m}\|\calA(\calX^{(t)} - \calX^\star) \|_F^2- \frac{1}{m}\<\calA^*(\calE), \calX^{(t)} - \calX^\star \>\nonumber\\
    &\!\!\!\!\geq\!\!\!\!& (1 - \delta_{2\wt r})\|\calX^{(t)} - \calX^\star\|_F^2  - \frac{\|\calX^{(t)} - \calX^\star\|_F}{m}\max_{\calH\in\setX_{2\ol r}, \|\calH\|_F\leq 1}\<\calA^*(\calE), \calH \>\nonumber\\
    &\!\!\!\!\geq\!\!\!\!& (1 - \delta_{2\wt r})\|\calX^{(t)} - \calX^\star\|_F^2 -\|\calX^{(t)} - \calX^\star\|_F  \cdot O\bigg(\frac{\ol r\gamma\sqrt{(1+\delta_{2\wt r})(N+M)\ol d(\log (N+M))}}{\sqrt{m}}\bigg),
\end{eqnarray}
where the last line follows the same result in \eqref{the tail function of fixed gaussian random variable 3} with $1-e^{-c_1N\wt d\wt r^2 \log N} -e^{-c_2(N+M)\ol d\ol r^2 \log (N+M)}$ for positive constants $c_1$ and $c_2$.

In addition, with probability $1-e^{-c_2(N+M)\ol d\ol r^2 \log (N+M)}-e^{-c_1N\wt d\wt r^2 \log N}$, we can derive
\begin{eqnarray}
    \label{upper bound of squared term in IHT l2}
    &\!\!\!\!\!\!\!\!&\|\nabla g(\calX^{(t)})\|_{F,2\ol r} - \|\calX^{(t)} - \calX^\star\|_{F,2\ol r}\nonumber\\
    &\!\!\!\!\leq \!\!\!\!& \|\nabla g(\calX^{(t)}) - \calX^{(t)} + \calX^\star\|_{F,2\ol r}\nonumber\\
    &\!\!\!\!=\!\!\!\!& \max_{\calZ\in\setX_{2\ol r}, \|\calZ\|_F\leq 1 } \bigg(  \frac{1}{m}\<\calA(\calX^{(t)}-\calX^\star)-\calE, \calA(\calZ)  \>    - \< \calX^{(t)}-\calX^\star,  \calZ\>  \bigg)\nonumber\\
    &\!\!\!\!\leq \!\!\!\!& \delta_{4 \wt r}\|\calX^{(t)}-\calX^\star\|_F  + O\bigg(\frac{\ol r\gamma\sqrt{(1+\delta_{2\wt r})(N+M)\ol d(\log (N+M))}}{\sqrt{m}}\bigg),
\end{eqnarray}
where the last line uses \Cref{RIP CONDITION FRO THE TENSOR TRAIN SENSING OTHER PROPERTY} and upper bound of $\max_{\calZ\in\setX_{2\ol r}, \|\calZ\|_F\leq 1 }\frac{1}{m}\<\calE,  \calA(\calZ)  \>$ in \eqref{the tail function of fixed gaussian random variable 3}.

Combing \eqref{lower bound of cross term in IHT l2} and \eqref{upper bound of squared term in IHT l2}, $\| {\widetilde{\calX}}^{(t)}   - \calX^\star \|_{F,2\ol r}^2$ can be rewritten as
\begin{eqnarray}
    \label{expansion of imtermediate term}
    &\!\!\!\!\!\!\!\!&\| {\widetilde{\calX}}^{(t)}   - \calX^\star \|_{F,2\ol r}^2\nonumber\\
     &\!\!\!\!\leq\!\!\!\!& (1 - 2\mu(1 - \delta_{4\wt r}) + 2\mu^2(1 + \delta_{4\wt r})^2  )\|\calX^{(t)} - \calX^\star\|_F^2 + (2\mu\|\calX^{(0)} - \calX^\star\|_F b + 2\mu^2 b^2 )\nonumber\\
    &\!\!\!\!\leq\!\!\!\!& (1 - \mu(1 - \delta_{4\wt r})   )\|\calX^{(t)} - \calX^\star\|_F^2  + (2\mu\|\calX^{(0)} - \calX^\star\|_F b + 2\mu^2 b^2 ),
\end{eqnarray}
where we define $b = O(\frac{\ol r\gamma\sqrt{(1+\delta_{2\wt r})(N+M)\ol d(\log (N+M))}}{\sqrt{m}})$ and the last line uses $\mu\leq \frac{1 - \delta_{4\wt r}}{2(1 + \delta_{4\wt r})^2}$.

Plugging \eqref{expansion of imtermediate term} into \eqref{The MSE relationship of different time in l2}, we can obtain
\begin{eqnarray*}
    \label{The expansion of difference two tensors in l2 2}
    \| \calX^{(t+1)}   - \calX^\star \|_F^2\leq a^{t+1}\|\calX^{(0)} - \calX^\star\|_F^2 +  O\bigg(\frac{1+ \frac{600(N+M)}{\underline{\sigma}({\calX^\star})}\|\calX^{(0)} - \calX^\star \|_F}{1-a}  (b\|\calX^{(0)} - \calX^\star\|_F + b^2) \bigg),
\end{eqnarray*}
where we require $\frac{\frac{600(N+M)}{\underline{\sigma}({\calX^\star})}\|\calX^{(0)} - \calX^\star \|_F}{(1+\frac{600(N+M)}{\underline{\sigma}({\calX^\star})}\|\calX^{(0)} - \calX^\star \|_F)(1-\delta_{4\wt r})}<\mu \leq \frac{1 - \delta_{4\wt r}}{2(1 + \delta_{4\wt r})^2}$ and $\|\calX^{(0)} - \calX^\star\|_F\leq \frac{(1 - \delta_{4\wt r})^2\underline{\sigma}({\calX^\star})}{600(N+M) (1 + \delta_{4\wt r}^2 + 6 \delta_{4\wt r})}$ to ensure that
$$a = \bigg( 1+ \frac{600(N+M)}{\underline{\sigma}({\calX^\star})}\|\calX^{(0)} - \calX^\star \|_F \bigg)(1 - \mu(1 - \delta_{4\wt r})   ) < 1. $$

\end{proof}

\section{Proof of \Cref{Local Convergence of Riemannian in the noisy ToT regression_Theorem calX}}
\label{proof of ToT factorization}

\subsection{Measure Distance Between Tensor Factors}

Before analyzing the Riemannian gradient descent algorithm, we will establish an error metric to quantify the distinctions between factors in two left-orthogonal form tensors, namely   $\calX = [{\mX}_1,\dots,{\mX}_{N+M} ]$ and  $\calX^\star = [{\mX}_1^\star,\dots,{\mX}_{N+M}^\star ]$.
Note that the left-orthogonal form still has rotation ambiguity among the factors in the sense that $\Pi_{i=1}^{N+M}\mX_i^\star(s_i) = \Pi_{i=1}^{N+M}\mR_{i-1}^\top\mX_i^\star(s_i)\mR_i$ for any orthonormal matrix $\mR_i\in\O^{r_i\times r_i}$ (with $\mR_0 = \mR_{N+M} = 1$). To capture this rotation ambiguity, by defining the rotated factors $L_{\mR}(\mX_i^\star)$ as
\begin{eqnarray}
\label{A modified left unfolding}
    L_{\mR}(\mX_i^\star) = \begin{bmatrix}\mR_{i-1}^\top\mX_{i}^\star(1)\mR_i\\ \vdots \\ \mR_{i-1}^\top\mX_{i}^\star(d_i)\mR_i\end{bmatrix}, 
\end{eqnarray}
we then define the distance between the two sets of factors as
\begin{eqnarray}
\label{BALANCED NEW DISTANCE BETWEEN TWO TENSORS}
    \text{dist}^2(\{\mX_i \},\{\mX_i^\star \}) = \min_{\mR_i\in\O^{r_i\times r_i}, \atop i \in [N + M -1]}\sum_{i=1}^{N + M -1} \ol\sigma^2(\calX^\star)\|L({\mX}_i)   -L_{\mR}({\mX}_i^\star)\|_F^2 + \|L({\mX}_{N+M})-L_{\mR}({\mX}_{N+M}^\star)\|_2^2,
\end{eqnarray}
where we note that $L({\mX}_1) = \mX_1$, $L_{\mR}({\mX}_1^\star) = \mX_1^\star\mR_1$ and $L({\mX}_{N+M}), L_{\mR}({\mX}_{N+M}^\star)\in\R^{(r_{N + M -1}d_{N+M})\times 1}$ are vectors.
Here coefficients $\ol\sigma^2(\calX^\star)$ and $1$ are incorporated to harmonize the energy between $\{L_{\mR}({\mX}_i^\star)  \}_{i\leq N + M -1}$ and $L_{\mR}({\mX}_{N+M}^\star)$ since $\|L_{\mR}({\mX}_i^\star)\|^2 = 1, i\in[N + M -1]$ and $\|L({\mX}_{N+M})-L_{\mR}({\mX}_{N+M}^\star)\|_2 = \|{\mX}_{N+M} - \mR_{N+M-1}^\top{\mX}_{N+M}^\star\|_F$, $\|\mX_{N+M}^\star\|^2  = \sigma_1^2({\calX^\star}^{\<N+M-1 \>}) \leq\ol\sigma^2(\calX^\star)$.
The following result establishes a connection between $\text{dist}^2(\{\mX_i \},\{ \mX_i^\star \})$ and $\|\calX-\calX^\star\|_F^2$.
\begin{lemma}(\cite[Lemma 1]{qin2024guaranteed})
\label{LOWER BOUND OF TWO DISTANCES main paper}
For any two left orthogonal TT formats $\calX = [\mX_1,\dots, \mX_{N+M}] $ and $\calX^\star = [\mX_1^\star,\dots, \mX_{N+M}^\star]$ with ranks $\vr = (r_1,\dots, r_{N + M -1})$ and $\ol{\sigma}^2(\calX)\leq \frac{9\ol{\sigma}^2(\calX^\star)}{4}$, the following holds:
\begin{eqnarray}
    \label{LOWER BOUND OF TWO DISTANCES_1 main paper}
    \hspace{-0.5cm}\|\calX-\calX^\star\|_F^2&\!\!\!\!\geq\!\!\!\!&\frac{\text{dist}^2(\{\mX_i \},\{\mX_i^\star \})}{8(N + M +1+\sum_{i=2}^{N + M -1}r_i)\kappa^2(\calX^\star)},\\
    \label{UPPER BOUND OF TWO DISTANCES_1 main paper}
    \hspace{-0.5cm}\|\calX-\calX^\star\|_F^2&\!\!\!\!\leq\!\!\!\!&\frac{9(N+M)}{4}\text{dist}^2(\{\mX_i \},\{\mX_i^\star \}),
\end{eqnarray}
where $\text{dist}^2(\{\mX_i \},\{\mX_i^\star \})$ is defined in \eqref{BALANCED NEW DISTANCE BETWEEN TWO TENSORS}.
\end{lemma}
\Cref{LOWER BOUND OF TWO DISTANCES main paper} ensures that $\calX$ is close to $\calX^\star$ once the corresponding factors are close with respect to the proposed distance measure, and the convergence behavior of $\|\calX-\calX^\star\|_F^2$ is reflected by the convergence in terms of the factors.

\subsection{Convergence Analysis with Respect to Tensor Factors}

Since RGD operates on the factors, we first examine the convergence behavior of RGD with respect to the tensor factors. We will then use \Cref{LOWER BOUND OF TWO DISTANCES main paper} to obtain the convergence for the entire tensor to prove \Cref{Local Convergence of Riemannian in the noisy ToT regression_Theorem calX}. Specifically,
the following result first ensures that, given an appropriate initialization, the RGD will converge to the target tensor up to a certain distance in \eqref{BALANCED NEW DISTANCE BETWEEN TWO TENSORS} that is proportional to the noise level.
\begin{theorem}
\label{Local Convergence of Riemannian in the noisy ToT regression_Theorem}[Convergence of tensor factors]
Denote the left orthogonal TT format of $\calX^\star$ as $[\mX_1^\star,\dots,\mX_{N+M}^\star]$ with ranks $\vr = (r_1,\dots, r_{N + M-1})$. Assume that $\calA$ obeys the $(N + M +3)\wt r$-RIP with a constant $\delta_{(N + M +3)\wt r}\leq\frac{7}{30}$, where $\wt r=\max_{i=1}^{N-1} r_i$. Suppose that the RGD in \eqref{RIEMANNIAN_GRADIENT_DESCENT ToT_1_1} and \eqref{RIEMANNIAN_GRADIENT_DESCENT ToT_1_2} is initialized with $\{\mX_i^{(0)} \}$ satisfying
\begin{eqnarray}
    \label{Local Convergence of Riemannian in the noisy ToT regression_Theorem initialization}
     \text{dist}^2(\{\mX_i^{(0)} \},\{\mX_i^\star \})\leq  \frac{(7 - 30\delta_{(N + M +3)\wt r})\underline{\sigma}^2(\calX^\star)}{8(N
    + M +1+  \sum_{i=2}^{N + M -1}  r_i)(129(N+M)^2+7231(N+M)-7360) }\nonumber
\end{eqnarray}
and uses the step size $\mu\leq\frac{7 - 30\delta_{(N + M +3)\wt r}}{20(9(N+M)-5)(1+\delta_{(N + M +3)\wt r})^2}$.  Then, with probability at least $1 - e^{-\Omega(N^3\wt d\wt r^2 \log N)} - (N+ M)e^{-\Omega(N\wt d\wt r^2 \log N)}  - (N+M) e^{-\Omega((N+M)\ol d\ol r^2 \log (N+M))}- e^{-\Omega((N+M)^3\ol d\ol r^2 \log (N+M))}$, the iterates $\{\mX_i^{(t)} \}_{t\geq 0}$ generated by the RGD satisfies
\begin{eqnarray}
    \label{Local Convergence of Riemannian in the noisy ToT regression_Theorem_1}
    \text{dist}^2(\{\mX_i^{(t+1)} \},\{\mX_i^\star \}) &\!\!\!\!\leq\!\!\!\!&\bigg(1-\frac{7 - 30\delta_{(N + M +3)\wt r}}{1280(N + M +1+\sum_{i=2}^{N + M -1}r_i)\kappa^2(\calX^\star)}\mu\bigg)^{t+1} \text{dist}^2(\{\mX_i^{(0)} \},\{\mX_i^\star \}) \nonumber\\
    &\!\!\!\!\!\!\!\!& + O\bigg(\frac{(1+\delta_{(N + M +3)\wt r}) (N+M)^4\ol d\ol r^3(\log (N+M)) \kappa^2(\calX^\star)\gamma^2 }{m(7 - 30\delta_{(N + M +3)\wt r})}\bigg)
\end{eqnarray}
as long as $m\geq C \frac{(N+M)^5\ol d \ol r^3 (\log (N+M)) \ol\sigma^2(\calX^\star) \gamma^2}{\underline{\sigma}^4(\calX^\star)} $ with a universal constant $C$, $\wt d=\max_{i=1}^{N} d_i$, $\ol d=\max_{i=1}^{N+M} d_i$ and $\ol r = \max_{i=1}^{N+M-1}r_i$.
\end{theorem}

\begin{proof}[Proof of \Cref{Local Convergence of Riemannian in the noisy ToT regression_Theorem}]
We first present useful properties for the factors $L(\mX_i^{(t)})$. Due to the retraction, $L(\mX_i^{(t)}), i\in[N + M -1]$ are always orthonormal. For $L(\mX_{N+M}^{(t)})$, assuming that \begin{align}
\text{dist}^2(\{\mX_i^{(t)} \},\{\mX_i^\star \})\leq \frac{\underline{\sigma}^2(\calX^\star)}{180(N+M)},
\label{eq:ini-cond-for-XN ToT}\end{align}
which is true for $t = 0$ and will be proved later for $t\ge 1$, then according to \cite[Appendix D]{qin2024guaranteed}
we obtain that $\ol{\sigma}^2(\calX^{(t)}) \leq \frac{9\ol{\sigma}^2(\calX^\star)}{4}$.

Utilizing the error metric defined in \eqref{BALANCED NEW DISTANCE BETWEEN TWO TENSORS}, we  define the best rotation matrices to align $\{\mX_i^{(t)} \}$ and $\{\mX_i^\star \}$ as
\begin{eqnarray}
    \label{the definition of orthonormal matrix R}
    (\mR_1^{(t)},\dots,\mR_{N + M -1}^{(t)}) =  \argmin_{\mR_i\in\O^{r_i\times r_i}, \atop i \in [N + M -1]}\sum_{i=1}^{N + M -1} \ol{\sigma}^2(\calX^\star)\|L({\mX}_i^{(t)})   -L_{\mR}({\mX}_i^\star)\|_F^2 + \|L(\mX_{N+ M}^{(t)})-L_{\mR}({\mX}_{N+ M}^\star)\|_2^2,
\end{eqnarray}
where $L_{\mR}({\mX}_i^\star)$ is defined in \eqref{A modified left unfolding}. We now expand $\text{dist}^2(\{\mX_i^{(t+1)} \},\{\mX_i^\star \})$ as
 \begin{eqnarray}
    \label{expansion of distance in ToT tensor regression}
  &\!\!\!\! \!\!\!\!&\text{dist}^2(\{\mX_i^{(t+1)} \},\{\mX_i^\star \})\nonumber\\
  &\!\!\!\! =\!\!\!\!&  \sum_{i=1}^{N + M -1} \ol{\sigma}^2(\calX^\star)\bigg\| L(\mX_i^{(t+1)})-L_{\mR^{(t+1)}}(\mX_i^\star) \bigg\|_F^2  +\bigg|\bigg|L(\mX_{N+ M }^{(t+1)})-L_{\mR^{(t+1)}}(\mX_{N+ M}^\star)\bigg|\bigg|_2^2  \nonumber\\
  &\!\!\!\!\leq\!\!\!\!&  \sum_{i=1}^{N + M -1} \ol{\sigma}^2(\calX^\star)\bigg\| L(\mX_i^{(t+1)})-L_{\mR^{(t)}}(\mX_i^\star) \bigg\|_F^2  +\bigg|\bigg|L(\mX_{N+ M }^{(t+1)})-L_{\mR^{(t)}}(\mX_{N+ M}^\star)\bigg|\bigg|_2^2\nonumber\\
  &\!\!\!\!\leq\!\!\!\!& \sum_{i=1}^{N + M -1} \ol{\sigma}^2(\calX^\star)\bigg\| L(\mX_i^{(t)})-L_{\mR^{(t)}}(\mX_i^\star) - \frac{\mu}{\ol{\sigma}^2(\calX^\star)}  \calP_{\text{T}_{L({\mX}_i)} \text{St}}\bigg(\nabla_{L({\mX}_{i})}f(\mX_1^{(t)}, \dots, \mX_{N+ M}^{(t)})\bigg)\bigg\|_F^2\nonumber\\
    &\!\!\!\!\!\!\!\!& +\bigg\|L(\mX_{N+ M}^{(t)})-L_{\mR^{(t)}}(\mX_{N+ M}^\star)-\mu\nabla_{L({\mX}_{N+M})}f(\mX_1^{(t)}, \dots, \mX_{N+ M}^{(t)}) \bigg\|_2^2\nonumber\\
    &\!\!\!\!=\!\!\!\!& \text{dist}^2(\{\mX_i^{(t)} \},\{\mX_i^\star \})-2\mu\sum_{i=1}^{N + M} \bigg\< L(\mX_i^{(t)})-L_{\mR^{(t)}}(\mX_i^\star), \calP_{\text{T}_{L({\mX}_i)} \text{St}}\bigg(\nabla_{L({\mX}_{i})}f(\mX_1^{(t)}, \dots, \mX_{N+ M}^{(t)})\bigg)\bigg\>\nonumber\\
    &\!\!\!\!\!\!\!\!& +\mu^2\bigg(\frac{1}{\ol{\sigma}^2(\calX^\star)}\sum_{i=1}^{N + M-1}\bigg\|\calP_{\text{T}_{L({\mX}_i)} \text{St}}\bigg(\nabla_{L({\mX}_{i})}f(\mX_1^{(t)}, \dots,  \mX_{N+ M}^{(t)})\bigg)\bigg\|_F^2+\bigg\|\nabla_{L({\mX}_{N + M })}f(\mX_1^{(t)}, \dots, \mX_{N+ M}^{(t)}) \bigg\|_2^2\bigg),
\end{eqnarray}
where the second inequality follows from the nonexpansiveness property described in \Cref{NONEXPANSIVENESS PROPERTY OF POLAR RETRACTION_1} of \Cref{Technical tools used in proofs}, and in the last line, to simplify the expression, we also define a projection operator for the last factor as $\calP_{\text{T}_{L({\mX}_{N+M})} \text{St}}=\calI$ such that $\calP_{\text{T}_{L({\mX}_{N+M})} \text{St}}(\nabla_{L({\mX}_{N + M})}f(\mX_1^{(t)}, \dots, \mX_{N+M}^{(t)})) = \nabla_{L({\mX}_{N + M})}f(\mX_1^{(t)},  \dots, \mX_{N+M}^{(t)})$.

Note that the gradient  $\nabla_{L({\mX}_{i})}f(\mX_1^{(t)}, \dots, \mX_{N+ M}^{(t)}) = L(\nabla_{{\mX}_{i}}f(\mX_1^{(t)}, \dots, \mX_{N+ M}^{(t)}))$ is given as follows:
\begin{eqnarray}
    \label{the defi of gradient}
    \nabla_{{\mX}_{i}}f(\mX_1^{(t)}, \dots, \mX_{N+ M}^{(t)}) =  \frac{1}{m}\calG_2,
\end{eqnarray}
with
\begin{eqnarray*}
\calG_1 &\!\!\!\!=\!\!\!\!&  \mX_{i-1}^{(t)}\times_{1,2}^{1,2}(\cdots\times_{1,2}^{1,2}(\mX_2^{(t)}\times_{1,2}^{1,2}(\mX_1^{(t)}\times_1^1 (\calB\times_{1}^1( \calA([\mX_1^{(t)},\dots, \mX_{N+ M}^{(t)}]) - \calY) ))) )\in\R^{r_{i-1}\times d_i\times\cdots\times d_{N+M}},\\
    \calG_2 &\!\!\!\!= \!\!\!\!& \calG_1 \times_{N+M-i+2}^2\mX_{N+M}^{(t)} \times_{N+M-i+1,N+M-i+2}^{2,3}\mX_{N+M-1}^{(t)} \times_{N+M-i,N+M-i+1}^{2,3}\cdots \times_{3,4}^{2,3}  \mX_{i+1}^{(t)}\in\R^{r_{i-1}\times d_i\times r_i},
\end{eqnarray*}
where $\calA([\mX_1, \dots, \mX_{N+ M}]) = \calB\times_2^1\mX_1\times_{N+1,2}^{1,2}\mX_2\times_{N,2}^{1,2}\mX_3\times_{N-1,2}^{1,2}\cdots\times_{3,2}^{1,2}
\mX_N\times_{2}^1\mX_{N+1}\times_3^1\mX_{N+2}\times_4^1\cdots\times_{M+1}^1\mX_{N+M}$ with $\mX_1$ and $\mX_{N+M}$ being reshaped as matrices of size $d_1\times r_1$ and $r_{N+M-1}\times d_{N+M}$, respectively.

The remainder of the proof consists of four components: (1) deriving an upper bound for the third term of \eqref{expansion of distance in ToT tensor regression}, (2) deriving a lower bound for the second term of \eqref{expansion of distance in ToT tensor regression}, (3) plugging these results into \eqref{expansion of distance in ToT tensor regression} to show the decay of the distance after one graduate update, and then obtaining
\eqref{Local Convergence of Riemannian in the noisy ToT regression_Theorem_1} by induction, and (4) finally proving \eqref{eq:ini-cond-for-XN ToT}, also by induction.

\paragraph*{Upper bound of the third term in \eqref{expansion of distance in ToT tensor regression}} We first define
\begin{eqnarray}
    \label{expansion of distance in ToT tensor factorization}
    &\!\!\!\!\!\!\!\!&\nabla_{{\mX}_{i}} F(\mX_1^{(t)},\dots,\mX_{N+M}^{(t)}))\nonumber\\
      &\!\!\!\!=\!\!\!\!&  (\mX_1^{(t)}\times_{2}^1 \cdots \times_{i-1}^1\mX_{i-1}^{(t)} )\times_{1,\dots,i-1}^{1,\dots,i-1} (\calX^{(t)} - \calX^\star) ) \times_{3,\dots,N+M-i+2}^{2,\dots, N+M-i+1}(\mX_{i+1}^{(t)} \times_3^1\cdots \times_{N+M-i+1}^1 \mX_{N+M}^{(t)} ),\nonumber
\end{eqnarray}
and then apply the \Cref{RIP CONDITION FRO THE TENSOR TRAIN SENSING OTHER PROPERTY} to obtain  the difference in the gradients of $f$
and $F$ through
\begin{eqnarray}
    \label{Riemannian GRADIENT DESCENT SQUARED TERM 1 to N+M ToT regression}
    &\!\!\!\!\!\!\!\!&\big\| \nabla_{L({\mX}_{i})} f(\mX_1^{(t)}, \dots, \mX_{N+M}^{(t)}) - \nabla_{L({\mX}_{i})} F(\mX_1^{(t)}, \dots, \mX_{N+M}^{(t)}) \big\|_F\nonumber\\
    &\!\!\!\!=\!\!\!\!& \max_{\mH_i\in\R^{r_{i-1}\times d_i\times r_i} \atop \|\mH_i\|_F\leq 1} \hspace{-0.15cm} \frac{1}{m}\<\calA(\calX^{(t)}-\calX^\star), \calA([\mX_1^{(t)},\dots, \mH_i, \dots,\mX_{N+M}^{(t)} ])   \>\nonumber\\
    &\!\!\!\!\!\!\!\!&- \<\calX^{(t)}-\calX^\star, [\mX_1^{(t)},\dots, \mH_i, \dots,\mX_{N+M}^{(t)} ]  \> - \frac{1}{m}\< \calA^* (\calE),  [\mX_1^{(t)},\dots, \mH_i, \dots,\mX_{N+M}^{(t)} ] \>    \nonumber\\
    &\!\!\!\!\leq\!\!\!\!& \delta_{3 \wt r}\|\calX^{(t)} - \calX^\star\|_F\|[\mX_1^{(t)},\dots, \mH_i, \dots,\mX_{N+M}^{(t)} ]\|_F  +\max_{\mH_i\in\R^{r_{i-1}\times d_i\times r_i} \atop \|\mH_i\|_F\leq 1}\frac{1}{m}\< \calA^* (\calE),  [\mX_1^{(t)},\dots, \mH_i, \dots,\mX_{N+M}^{(t)} ] \> \nonumber\\
    &\!\!\!\!\leq\!\!\!\!&\begin{cases}
    O\bigg(\frac{\ol r\sqrt{(1+\delta_{3 \wt r})(N+M)\ol d(\log (N+M)) }\gamma \ol{\sigma}(\calX^\star)}{\sqrt{m}}\bigg)
     +\frac{3\ol{\sigma}(\calX^\star)}{2}\delta_{3 \wt r}\|\calX^{(t)}-\calX^\star\|_F, &   i \in [N + M -1],\\
    O\bigg(\frac{\ol r\sqrt{(1+\delta_{3 \wt r})(N+M)\ol d(\log (N+M)) }\gamma}{\sqrt{m}}\bigg)
     +\delta_{3 \wt r}\|\calX^{(t)}-\calX^\star\|_F, &  i = N + M,
  \end{cases}
\end{eqnarray}
where $\mH_1\in\R^{d_1\times r_1}$ and $\mX_{N+M}\in\R^{r_{N+M-1}\times d_{N+M}}$.
The first inequality follows \eqref{RIP CONDITION FRO THE TENSOR TRAIN SENSING OTHER PROPERTY_1} in \Cref{RIP CONDITION FRO THE TENSOR TRAIN SENSING OTHER PROPERTY}. Following the same derivation of \eqref{the tail function of fixed gaussian random variable 3}, the second inequality holds with probability at least $1 - (N+M)e^{-\Omega(N\wt d\wt r^2 \log N)} - (N+M)e^{-\Omega((N+M)\ol d\ol r^2 \log (N+M))}$, and is derived by using  $\|[\mX_1^{(t)},\dots,  \mH_i, \dots,\mX_{N+M}^{(t)} ]\|_F\leq \|{\calX^{(t)}}^{\geq i+1}\| \leq \frac{3}{2}\ol{\sigma}(\calX^\star),  i\in[N + M -1]$ and $\|[\mX_1^{(t)}, \dots,  \mX_{N + M -1}^{(t)}, \mH_{N+M} ]\|_F =\|\mH_{N+M}\|_F\leq 1$.

Combining \eqref{Riemannian GRADIENT DESCENT SQUARED TERM 1 to N+M ToT regression} with
\begin{eqnarray}
    \label{SQUARED TERM 1 to N+M ToT F}
    &\!\!\!\!\!\!\!\!&\big\|  \nabla_{L({\mX}_{i})} F(\mX_1^{(t)}, \dots, \mX_{N+M}^{(t)})   \big\|_F \nonumber\\
    &\!\!\!\! = \!\!\!\!& \hspace{-0.2cm}\max_{\mH_i\in\R^{r_{i-1}\times d_i\times r_i} \atop \|\mH_i\|_F\leq 1} \<\calX^{(t)}-\calX^\star,  [\mX_1^{(t)},\dots, \mH_i, \dots,\mX_{N+M}^{(t)} ] \>\nonumber\\
    &\!\!\!\!\leq\!\!\!\!&\begin{cases}
    \frac{3\ol{\sigma}(\calX^\star)}{2}\|\calX^{(t)}-\calX^\star\|_F, & i \in [N + M -1], \\
    \|\calX^{(t)}-\calX^\star\|_F, & i = N + M,
    \end{cases}
\end{eqnarray}
we can obtain the following upper bound
\begin{eqnarray*}
    \label{Riemannian GRADIENT DESCENT SQUARED TERM 1 to N+M ToT regression1}
    &\!\!\!\!\!\!\!\!&\big\| \nabla_{L({\mX}_{i})} f(\mX_1^{(t)}, \dots, \mX_{N+M}^{(t)}) \big\|_F\nonumber\\
    &\!\!\!\!\leq\!\!\!\!&\begin{cases}
    O\bigg(\frac{\ol r\sqrt{(1+\delta_{3 \wt r})(N+M)\ol d(\log (N+M)) }\gamma \ol{\sigma}(\calX^\star)}{\sqrt{m}}\bigg)
    +\frac{3\ol{\sigma}(\calX^\star)}{2}(1 + \delta_{3 \wt r})\|\calX^{(t)}-\calX^\star\|_F, &  i \in [N + M -1],\\
    O\bigg(\frac{\ol r\sqrt{(1+\delta_{3 \wt r})(N+M)\ol d(\log (N+M)) }\gamma}{\sqrt{m}}\bigg)
     +(1 + \delta_{3 \wt r})\|\calX^{(t)}-\calX^\star\|_F, &  i = N + M.
  \end{cases}
\end{eqnarray*}

We now plug the above into the third term in  \eqref{expansion of distance in ToT tensor regression} to get
\begin{eqnarray}
    \label{upper bound of third term squared ToT}
    &\!\!\!\!\!\!\!\!&\hspace{-0.3cm}\frac{1}{\ol{\sigma}^2(\calX^\star)}\hspace{-0.3cm}\sum_{i=1}^{N + M-1}\big\|\calP_{\text{T}_{L({\mX}_i)} \text{St}}\big(\nabla_{L({\mX}_{i})} f(\mX_1^{(t)}, \dots, \mX_{N+M}^{(t)})\big)\big\|_F^2 +\big\|\nabla_{L({\mX}_{N +M})} f(\mX_1^{(t)}, \dots, \mX_{N+M}^{(t)}) \big\|_2^2\nonumber\\
    &\!\!\!\!\leq\!\!\!\!&\frac{1}{\ol{\sigma}^2(\calX^\star)}\sum_{i=1}^{N + M -1}\big\|\nabla_{L({\mX}_{i})} f(\mX_1^{(t)}, \dots, \mX_{N+M}^{(t)})\big\|_F^2+\big\|\nabla_{L({\mX}_{N + M})} f(\mX_1^{(t)}, \dots, \mX_{N+M}^{(t)}) \big\|_2^2\nonumber\\
    &\!\!\!\!\leq\!\!\!\!&\frac{9(N+M)-5}{2}(1+\delta_{3 \wt r})^2\|\calX^{(t)} - \calX^\star\|_F^2 + O\bigg(\frac{(1+\delta_{3 \wt r})(N+M)^2\ol d\ol r^2(\log (N+M)) \gamma^2 }{m}\bigg),
\end{eqnarray}
where the first inequality follows from the fact that for any matrix $\mB = \calP_{\text{T}_{L({\mX}_i)} \text{St}}(\mB) + \calP_{\text{T}_{L({\mX}_{i})} \text{St}}^{\perp}(\mB)$ where $\calP_{\text{T}_{L({\mX}_i)} \text{St}}(\mB)$ and $\calP_{\text{T}_{L({\mX}_{i})} \text{St}}^{\perp}(\mB)$ are orthogonal,  we have $\|\calP_{\text{T}_{L({\mX}_i)} \text{St}}(\mB)\|_F^2\leq \|\mB\|_F^2$.

\paragraph*{Lower bound of the second term in \eqref{expansion of distance in ToT tensor regression}} We first expand the second term in \eqref{expansion of distance in ToT tensor regression} as following:
\begin{eqnarray}
    \label{CROSS TERM LOWER BOUND_in ToT regression}
    &\!\!\!\!\!\!\!\!&\sum_{i=1}^{N + M} \big\< L(\mX_i^{(t)})-L_{\mR^{(t)}}(\mX_i^\star),\calP_{\text{T}_{L({\mX}_i)} \text{St}}\big(\nabla_{L({\mX}_{i})}f(\mX_1^{(t)},\dots, \mX_{N + M}^{(t)})\big)\big\>\nonumber\\
    &\!\!\!\!=\!\!\!\!& \sum_{i=1}^{N + M} \big\< L(\mX_i^{(t)})-L_{\mR^{(t)}}(\mX_i^\star), \nabla_{L({\mX}_{i})}f(\mX_1^{(t)},\dots, \mX_{N + M}^{(t)})\big\> - T\nonumber\\
    &\!\!\!\!=\!\!\!\!&\frac{1}{m}\big\<\calX^{(t)} - \calX^\star + \calH^{(t)}, \calB\times_1^1(\calA(\calX^{(t)}) - \calY ) \big\>-T\nonumber\\
    &\!\!\!\!=\!\!\!\!& \frac{1}{m}\|\calA(\calX^{(t)} - \calX^\star)\|_F^2 + \frac{1}{m}\<\calA(\calX^{(t)} - \calX^\star), \calA(\calH^{(t)})  \> - \frac{1}{m}\< \calX^{(t)} - \calX^\star + \calH^{(t)}, \calB\times_1^2 \calE \> - T,
\end{eqnarray}
where we define
\begin{eqnarray}
    \label{The definition of Ht ToT}
    \calH^{(t)} &\!\!\!\!=\!\!\!\!& \calX^\star - [\mX_1^{(t)},\dots,\mX_{N+M-1}^{(t)}, \mR_{N+M-1}^{(t)}\times_2^1 \mX_{N+M}^\star ]\nonumber\\
     &\!\!\!\!\!\!\!\!&+\sum_{i=1}^{N+M-1}[\mX_1^{(t)}, \dots, \mX_{i-1}^{(t)}, \mX_i^{(t)} - \mR_{i-1}^{(t)}\times_2^1\mX_i^\star  \times_3^1\mR_{i}^{(t)}, \mX_{i+1}^{(t)}, \dots, \mX_{N+M}^{(t)} ],
\end{eqnarray}
and
\begin{eqnarray}
    \label{The definition of T ToT}
    T  =  \sum_{i=1}^{N + M -1} \big\<\calP^{\perp}_{\text{T}_{L({\mX}_i)} \text{St}}(L(\mX_i^{(t)})-L_{\mR^{(t)}}(\mX_i^\star)),  \nabla_{L({\mX}_{i})}f(\mX_1^{(t)}, \dots, \mX_{N+M}^{(t)}) \big\>.
\end{eqnarray}
Here $\calP^{\perp}_{\text{T}_{L({\mX}_i)} \text{St}}(L(\mX_i^{(t)})-L_{\mR^{(t)}}(\mX_i^\star)) = L(\mX_i^{(t)})-L_{\mR^{(t)}}(\mX_i^\star)   - \calP_{\text{T}_{L({\mX}_i)} \text{St}}(L(\mX_i^{(t)})-L_{\mR^{(t)}}(\mX_i^\star))$. According to \cite[eqs. (82) \text{and} (119)]{qin2024guaranteed}, we can respectively obtain
\begin{align}
    \label{Upper bound of H in  ToT factorization}
    \|\calH^{(t)}\|_F^2\leq \frac{9(N+M)(N +M-1)}{8\ol{\sigma}^2(\calX^\star)}\text{dist}^4(\{\mX_i^{(t)} \},\{\mX_i^\star \}),
\end{align}
and
\begin{eqnarray*}
    \label{Upper bound of T in  ToT factorization}
    \hspace{-0.5cm}T \leq  \frac{1}{20}\|\calX^{(t)}-\calX^\star\|_F^2+\frac{46(N + M -1)}{\ol{\sigma}^2(\calX^\star)}\text{dist}^4(\{\mX_i^{(t)} \},\{\mX_i^\star \})  + O\bigg(\frac{(1+\delta_{3 \wt r}) (N+ M)^2\ol d\ol r^2(\log (N+ M))\gamma^2 }{m}\bigg).
\end{eqnarray*}
In addition, from \cite[eq. (120)]{qin2024guaranteed}, with probability $1 - e^{-\Omega((N+M)^3\ol d\ol r^2 \log (N+M))} - e^{-\Omega(N^3\wt d\wt r^2 \log N)}$, we have
\begin{eqnarray}
    \label{Upper bound of error term in  ToT factorization}
\frac{1}{m}\< \calX^{(t)} - \calX^\star + \calH^{(t)}, \calB\times_1^2 \calE \>&\!\!\!\!\leq\!\!\!\!&\frac{1}{10}\|\calX^{(t)} - \calX^\star\|_F^2 + O\bigg(\frac{(1+\delta_{(N + M+3)\ol r})(N + M)^3\ol d\ol r^2(\log (N + M)) \gamma^2}{m}\bigg)\nonumber\\
    &\!\!\!\! \!\!\!\!&+ \frac{9(N + M)(N + M -1)}{80\ol{\sigma}^2(\calX^\star)}\text{dist}^4(\{\mX_i^{(t)} \},\{\mX_i^\star \}).
\end{eqnarray}
Now we plug them into \eqref{CROSS TERM LOWER BOUND_in ToT regression} to get
\begin{eqnarray}
    \label{CROSS TERM LOWER BOUND_in ToT regression1}
    &\!\!\!\!\!\!\!\!&\sum_{i=1}^{N + M} \big\< L(\mX_i^{(t)})-L_{\mR^{(t)}}(\mX_i^\star),\calP_{\text{T}_{L({\mX}_i)} \text{St}}\big(\nabla_{L({\mX}_{i})}f(\mX_1^{(t)},\dots, \mX_{N + M}^{(t)})\big)\big\> \nonumber\\
    &\!\!\!\!\geq\!\!\!\!&(1-\delta_{2\wt r})\|\calX^{(t)}-\calX^\star\|_F^2 + \< \calX^{(t)} - \calX^\star, \calH^{(t)} \>- \delta_{(N+M+3)\wt r} \|\calX^{(t)} - \calX^\star\|_F \|\calH^{(t)}\|_F\nonumber\\
    &\!\!\!\! \!\!\!\!&- \frac{1}{m}\< \calX^{(t)} - \calX^\star + \calH^{(t)}, \calB\times_1^2 \calE \> - T\nonumber\\
    &\!\!\!\!\geq\!\!\!\!&\big(\frac{9}{20}-\frac{3\delta_{(N + M +3)\wt r}}{2}\big)\|\calX^{(t)}-\calX^\star\|_F^2- \frac{1+\delta_{(N + M +3)\wt r}}{2}\|\calH^{(t)}\|_F^2 -\frac{46(N + M -1)}{\ol{\sigma}^2(\calX^\star)}\text{dist}^4(\{\mX_i^{(t)} \},\{\mX_i^\star \})\nonumber\\
    &\!\!\!\!\!\!\!\!&-\frac{1}{m}\< \calX^{(t)} - \calX^\star + \calH^{(t)}, \calB\times_1^2 \calE \> - O\bigg(\frac{(1+\delta_{3 \wt r}) (N+M)^2\ol d\ol r^2(\log (N+M))\gamma^2 }{m}\bigg)\nonumber\\
    &\!\!\!\!\geq\!\!\!\!&\frac{7 - 30\delta_{(N + M +3)\wt r}}{40}\|\calX^{(t)}-\calX^\star\|_F^2  -\frac{129(N + M)^2+7231(N + M)-7360}{160\ol{\sigma}^2(\calX^\star)}\text{dist}^4(\{\mX_i^{(t)} \},\{\mX_i^\star \})\nonumber\\
    &\!\!\!\!\!\!\!\!& - O\bigg(\frac{(1+\delta_{(N + M +3)\wt r}) (N + M)^3\ol d\ol r^2(\log (N + M))\gamma^2 }{m}\bigg)\nonumber\\
    &\!\!\!\!\geq\!\!\!\!& \frac{(7 - 30\delta_{(N + M +3)\wt r})\underline{\sigma}^2(\calX^\star)}{1280(N + M +1+\sum_{i=2}^{N + M -1}r_i)\ol{\sigma}^2(\calX^\star)} \text{dist}^2(\{\mX_i^{(t)} \},\{\mX_i^\star \})+ \frac{7 - 30\delta_{(N + M +3)\wt r}}{80}\|\calX^{(t)}-\calX^\star\|_F^2\nonumber\\
    &\!\!\!\!\!\!\!\!& - O\bigg(\frac{(1+\delta_{(N + M +3)\wt r}) (N+M)^3\ol d\ol r^2(\log (N+M))\gamma^2 }{m}\bigg),
\end{eqnarray}
where we apply \Cref{RIP condition fro the ToT TT regression Lemma} and \Cref{RIP CONDITION FRO THE TENSOR TRAIN SENSING OTHER PROPERTY} in the first inequality, and use $\delta_{(N + M+3)\wt r}\leq\frac{7}{30}$, \eqref{Upper bound of H in  ToT factorization} and \eqref{Upper bound of error term in  ToT factorization} in the second inequality. The last line follows \Cref{LOWER BOUND OF TWO DISTANCES main paper} and the initial condition $\text{dist}^2(\{\mX_i^{(0)} \},\{\mX_i^\star \})\leq (7 - 30\delta_{(N + M +3)\wt r}) \underline{\sigma}^2(\calX^\star)/\big(8(N + M +1+  \sum_{i=2}^{N + M -1} r_i)(129(N+M)^2+7231(N+M)-7360) \big)$.

\paragraph*{Proof of \eqref{Local Convergence of Riemannian in the noisy ToT regression_Theorem_1} by induction}

Taking \eqref{upper bound of third term squared ToT} and \eqref{CROSS TERM LOWER BOUND_in ToT regression1} into \eqref{expansion of distance in ToT tensor regression}, with probability $1 - (N+M)e^{-\Omega((N+M)\ol d\ol r^2 \log (N+M))} - (N+M)e^{-\Omega(N\wt d\wt r^2 \log N)}- e^{-\Omega((N+M)^3\ol d\ol r^2 \log (N+M))} - e^{-\Omega(N^3\wt d\wt r^2 \log N)}$, we can get 
\begin{eqnarray}
    \label{Conclusion of Riemannian gradient descent in the noisy ToT regression}
    \text{dist}^2(\{\mX_i^{(t+1)} \},\{\mX_i^\star \})&\!\!\!\!\leq\!\!\!\!&\bigg(1-\frac{(7 - 30\delta_{(N + M +3)\wt r})\underline{\sigma}^2(\calX^\star)}{1280(N + M +1+\sum_{i=2}^{N + M -1}r_i)\ol{\sigma}^2(\calX^\star)}\mu\bigg)\text{dist}^2(\{\mX_i^{(t)} \},\{\mX_i^\star \})\nonumber\\
    &\!\!\!\!\!\!\!\!&+\bigg(\frac{9(N+M)-5}{2}(1+\delta_{3 \wt r})^2\mu^2  -\frac{7 - 30\delta_{(N + M +3)\wt r}}{40}\mu \bigg)\|\calX^{(t)}-\calX^\star\|_F^2\nonumber\\
    &\!\!\!\!\!\!\!\!&+ O\bigg(\frac{(1+\delta_{(N + M +3)\wt r})(N+M)^2\ol d\ol r^2(\log (N+M)) \gamma^2}{m}(\mu (N+M) + \mu^2)\bigg)\nonumber\\
    &\!\!\!\!\leq\!\!\!\!&\bigg(1-\frac{(7 - 30\delta_{(N + M+3)\wt r})\underline{\sigma}^2(\calX^\star)}{1280(N + M +1+\sum_{i=2}^{N + M -1}r_i)\ol{\sigma}^2(\calX^\star)}\mu\bigg)\text{dist}^2(\{\mX_i^{(t)} \},\{\mX_i^\star \})\nonumber\\
    &\!\!\!\!\!\!\!\!&+ O\bigg(\frac{(1+\delta_{(N + M +3)\wt r})(N + M)^2\ol d\ol r^2(\log (N+ M)) \gamma^2}{m}(\mu (N+ M) + \mu^2)\bigg),
\end{eqnarray}
where we use $\mu\leq\frac{7 - 30\delta_{(N + M +3)\wt r}}{20(9(N+M)-5)(1+\delta_{(N + M +3)\wt r})^2}$ in the last line. By the induction, this further implies 
\begin{eqnarray}
    \label{Simiplified Conclusion of Riemannian gradient descent in the noisy ToT regression}
    &\!\!\!\!\!\!\!\!&\text{dist}^2(\{\mX_i^{(t+1)} \},\{\mX_i^\star \})\leq\bigg(1-\frac{(7 - 30\delta_{(N + M +3)\wt r})\underline{\sigma}^2(\calX^\star)}{1280(N + M +1+\sum_{i=2}^{N + M -1}r_i)\ol{\sigma}^2(\calX^\star)}\mu\bigg)^{t+1}
    \text{dist}^2(\{\mX_i^{(0)} \},\{\mX_i^\star \})\nonumber\\
    &\!\!\!\!\!\!\!\!&+  O\bigg(\frac{(N + M + \mu)(N + M +1+\sum_{i=2}^{N + M -1}r_i)(1+\delta_{(N + M +3)\wt r}) (N+ M)^2\ol d\ol r^2(\log (N+ M)) \ol{\sigma}^2(\calX^\star)\gamma^2 }{m(7 - 30\delta_{(N + M +3)\wt r})\underline{\sigma}^2(\calX^\star)}\bigg),
\end{eqnarray}
which can be simplified to \eqref{Local Convergence of Riemannian in the noisy ToT regression_Theorem_1}.

\paragraph*{Proof of \eqref{eq:ini-cond-for-XN ToT}} We can now prove \eqref{eq:ini-cond-for-XN ToT} by induction. First note that \eqref{eq:ini-cond-for-XN ToT} holds for $t = 0$. We now assume it holds for all $t \le t'$, which implies that $\sigma_1^2({\calX^{(t')}}^{\<i \>}) =   \|{\calX^{(t')}}^{\geq i+1} \|^2\leq \frac{9\ol{\sigma}^2(\calX^\star)}{4}, i\in[N + M -1]$. By invoking \eqref{Simiplified Conclusion of Riemannian gradient descent in the noisy ToT regression}, we have
\begin{align*}
    \text{dist}^2( \{\mX_i^{(t'+1)}\},\{\mX_i^\star\})& \le \text{dist}^2(\{\mX_i^{(0)} \},\{\mX_i^\star \}) +  O\big(\frac{(1+\delta_{(N + M +3)\wt r}) (N+M)^4\ol d\ol r^3(\log (N+M)) \ol{\sigma}^2(\calX^\star)\gamma^2 }{m(7 - 30\delta_{(N + M +3)\wt r})\underline{\sigma}^2(\calX^\star)}\big)\nonumber\\
   & \le \frac{\underline{\sigma}^2(\calX^\star)}{180(N+M)},
\end{align*}
as long as $
    m\geq C \frac{(N+M)^5\ol d \ol r^3 (\log (N+M))\ol{\sigma}^2(\calX^\star) \gamma^2}{\underline{\sigma}^4(\calX^\star)} $ with a universal constant $C$. Consequently, \eqref{eq:ini-cond-for-XN ToT} also holds  at $ t= t'+1$. By induction, we can conclude that \eqref{eq:ini-cond-for-XN ToT} holds for all $t\ge 0$. This completes the proof.

\end{proof}

\subsection{Convergence Analysis with Respect to Entire Tensor in \Cref{Local Convergence of Riemannian in the noisy ToT regression_Theorem calX}}

\begin{proof}
 Upon establishing \Cref{Local Convergence of Riemannian in the noisy ToT regression_Theorem}, we combine it together  with \Cref{LOWER BOUND OF TWO DISTANCES main paper} to conclude \Cref{Local Convergence of Riemannian in the noisy ToT regression_Theorem calX}.
\end{proof}

\section{Proof of \Cref{TENSOR SENSING SPECTRAL INITIALIZATION ToT}}
\label{proof of spectral init ToT}

Based on the definition in \eqref{Definition of the restricted F norm}, we can expand the expression for $\|\calX^{(0)}-\calX^\star\|_F$ as
\begin{eqnarray}
    \label{The expansion of noisy spectral ini L2}
\|\calX^{(0)}-\calX^\star\|_F&\!\!\!\!=\!\!\!\!&\bigg\|\text{SVD}_{\vr}^{tt}(\frac{1}{m}\calA^*(\calY))-\calX^\star\bigg\|_{F,2\ol r} \leq\!\bigg\|\text{SVD}_{\vr}^{tt}(\frac{1}{m}\calA^*(\calY)) \!-\! \frac{1}{m}\calA^*(\calY)\bigg\|_{\!F,2\ol r}
    \hspace{-0.3cm}+\bigg\|\frac{1}{m}\calA^*(\calY)-\calX^\star\bigg\|_{F,2\ol r}\nonumber\\
    &\!\!\!\!\leq\!\!\!\!&\sqrt{N+M-1}\bigg\|\text{opt}_{\vr}(\frac{1}{m}\calA^*(\calY))-\frac{1}{m}\calA^*(\calY)\bigg\|_{F,2\ol r}  + \bigg\|\frac{1}{m}\calA^*(\calY)-\calX^\star\bigg\|_{F,2\ol r}\nonumber\\
    &\!\!\!\!\leq\!\!\!\!&(1+\sqrt{N+M-1})\bigg\|\frac{1}{m}\calA^*(\calY)-\calX^\star\bigg\|_{F,2\ol r}\nonumber\\
    &\!\!\!\!\leq\!\!\!\!&(1+\sqrt{N+M-1})\max_{\calZ\in\setX_{2\ol r}, \|\calZ\|_F\leq 1 }\bigg(\frac{1}{m}\<\calA(\calX^\star),  \calA(\calZ)  \>  - \<\calX^\star, \calZ  \>  \bigg)+ (1+\sqrt{N+M-1})\bigg\|\frac{1}{m}\calA^*(\calE) \bigg\|_{F,2\ol r}\nonumber\\
    &\!\!\!\!\leq\!\!\!\!& (1+\sqrt{N+M-1}) \bigg(\delta_{3 \wt r}\|\calX^\star\|_F + O\bigg( \frac{\ol r\gamma\sqrt{(1+\delta_{2\wt r})(N+M)\ol d(\log (N+M))}}{\sqrt{m}}  \bigg)   \bigg),\nonumber
\end{eqnarray}
where $\text{opt}_{\vr}(\frac{1}{m}\calA^*(\calY))$ is the best TT-approximation of ranks $\vr$ to $\frac{1}{m}\calA^*(\calY)$ in the Frobenius norm, the second inequality utilizes the quasi-optimality property of TT-SVD projection \cite{Oseledets11}, the third inequality follows because the definition of $\text{opt}_{\vr}(\cdot)$ and $\calX^\star$ has ranks $\vr$, and the last uses \eqref{the tail function of fixed gaussian random variable 3} with probability $1-e^{-\Omega(N\wt d\wt r^2 \log N)} -e^{-\Omega(((N+M)\ol d\ol r^2 \log (N+M))}$.

\section{Technical tools used in proofs}
\label{Technical tools used in proofs}

\begin{lemma}(\cite[Lemma 10]{qin2024quantum})
\label{EXPANSION_A1TOAN-B1TOBN_1}
For any  ${\bm A}_i,{\bm A}^\star_i\in\R^{r_{i-1}\times r_i},i=1,\dots,N$, we have
\begin{eqnarray}
    \label{EXPANSION_A1TOAN-B1TOBN_2}
    {\bm A}_1{\bm A}_2\cdots {\bm A}_N-{\bm A}_1^\star{\bm A}_2^\star\cdots {\bm A}_N^\star =  \sum_{i=1}^N \mA_1^\star \cdots \mA_{i-1}^\star (\mA_{i} - \mA_i^\star) \mA_{i+1} \cdots \mA_N.
\end{eqnarray}

\end{lemma}

As an immediate consequence of the RIP, the inner product between two low-rank TT formats is also nearly preserved if $\calA$ satisfies the RIP.
\begin{lemma} (\cite{CandsTIT11,Rauhut17})
\label{RIP CONDITION FRO THE TENSOR TRAIN SENSING OTHER PROPERTY}
Suppose that $\calA$ obeys the $2\wt r$-RIP with a constant $\delta_{2\wt r}\in(0,1)$. Then for any TT formats $\calX_1,\calX_2\in\R^{d_{1} \times \cdots \times d_{N+M}}$ of rank at most $\wt r$, one has
\begin{eqnarray}
    \label{RIP CONDITION FRO THE TENSOR TRAIN SENSING OTHER PROPERTY_1}
    \bigg|\frac{1}{m}\<\calA(\calX_1),\calA(\calX_2)\>-\<\calX_1,\calX_2\>\bigg|\leq \delta_{2\wt r}\|\calX_1\|_F\|\calX_2\|_F.
\end{eqnarray}
\end{lemma}

\begin{lemma}(\cite[Lemma 26]{cai2022provable})
\label{Perturbation bound for TT SVD}
Let $\calX^\star$ be in TT format with the ranks $(r_1,\dots, r_{N-1})$. Given  a noisy tensor $\calX=\calX^\star+\mathcal{D}$ such that $ C_N ||\mathcal{D}||_F  \leq \underline{\sigma}({\calX^\star})$ for some constant $C_N\geq 500 N$, we have
\begin{eqnarray}
    \label{Perturbation bound for TT SVD1}
    ||\text{SVD}_{\vr}^{tt}(\calX)-\calX^\star||_F^2\leq||\mathcal{D}||_F^2+\frac{600N||\mathcal{D}||_F^3}{\underline{\sigma}({\calX^\star})},
\end{eqnarray}
where $\text{SVD}_{\vr}^{tt}(\cdot)$ represents the TT-SVD operation.
\end{lemma}

\begin{lemma}(\cite[Lemma 1]{LiSIAM21})
\label{NONEXPANSIVENESS PROPERTY OF POLAR RETRACTION_1}
Let ${\mX}\in\text{St}(n,r)$ and ${\bm \xi}\in\text{T}_{\mX} \text{St}$ be given. Consider the point ${\mX}^+={\mX}+{\bm \xi}$. Then, the polar decomposition-based retraction $\text{Retr}_{\mX}({\mX}^+)={\mX}^+({{\mX}^+}^\top{\mX}^+)^{-\frac{1}{2}}$ satisfies
\begin{eqnarray}
    \label{NONEXPANSIVENESS PROPERTY OF POLAR RETRACTION_2}
    \|\text{Retr}_{\mX}(\mX^+)-\overline{\mX}\|_F&\!\!\!\! \leq \!\!\!\!&\|{\mX}^+-\overline{\mX}\|_F=\|{\mX}+{\bm \xi}-\overline{\mX}\|_F,  \  \forall\overline{\mX}\in\text{St}(n,r).
\end{eqnarray}
\end{lemma}

\vfill

\end{document}

%% file: macro.tex
\usepackage{url}
\usepackage[hidelinks]{hyperref}
\usepackage{amsmath,amsthm,amssymb,amsbsy}
\usepackage{paralist}
\usepackage{xcolor}
\usepackage{color}
\usepackage{comment}
\usepackage{multirow}

\usepackage{fancyhdr}
\usepackage{cite}
\usepackage{cleveref}

\newtheorem{lemma}{Lemma}

\newtheorem{theorem}{Theorem}

\theoremstyle{remark}

\theoremstyle{problem}

\renewcommand{\O}{\mathbb{O}}
\newcommand{\R}{\mathbb{R}}
\def \real    { \mathbb{R} }

\newcommand{\e}{\begin{equation}}
\newcommand{\ee}{\end{equation}}
\newcommand{\en}{\begin{equation*}}
\newcommand{\een}{\end{equation*}}
\newcommand{\eqn}{\begin{eqnarray}}
\newcommand{\eeqn}{\end{eqnarray}}
\newcommand{\bmat}{\begin{bmatrix}}
\newcommand{\emat}{\end{bmatrix}}

\DeclareMathAlphabet\mathbfcal{OMS}{cmsy}{b}{n}
\renewcommand{\P}[1]{\operatorname{\mathbb{P}}\left(#1\right)}
\newcommand{\E}{\operatorname{\mathbb{E}}}

\newcommand{\vct}[1]{\boldsymbol{#1}}
\newcommand{\mtx}[1]{\boldsymbol{#1}}

\newcommand{\<}{\langle}
\renewcommand{\>}{\rangle}

\newcommand{\set}[1]{\mathbb{#1}}

\DeclareMathOperator*{\argmin}{\text{arg~min}}
\DeclareMathOperator*{\argmax}{\text{arg~max}}
\def \st {\operatorname*{s.t.\ }}

\newcommand{\wh}{\widehat}

\newcommand{\wt}{\widetilde}
\newcommand{\ol}{\overline}

\newcommand{\calA}{\mathcal{A}}
\newcommand{\calB}{\mathcal{B}}

\newcommand{\calE}{\mathcal{E}}

\newcommand{\calG}{\mathcal{G}}
\newcommand{\calH}{\mathcal{H}}
\newcommand{\calI}{\mathcal{I}}

\newcommand{\calN}{\mathcal{N}}

\newcommand{\calP}{\mathcal{P}}

\newcommand{\calX}{\mathcal{X}}
\newcommand{\calY}{\mathcal{Y}}
\newcommand{\calZ}{\mathcal{Z}}

\newcommand{\vr}{\vct{r}}

\newcommand{\vy}{\vct{y}}

\newcommand{\mA}{\mtx{A}}
\newcommand{\mB}{\mtx{B}}

\newcommand{\mH}{\mtx{H}}

\newcommand{\mP}{\mtx{P}}

\newcommand{\mR}{\mtx{R}}

\newcommand{\mV}{\mtx{V}}

\newcommand{\mX}{\mtx{X}}
\newcommand{\mY}{\mtx{Y}}

\newcommand{\mId}{{\bf I}}

\newcommand{\setX}{\set{X}}

\graphicspath{{./figs/}}

\newlength{\imgwidth}
\newboolean{twoColVersion}
\setboolean{twoColVersion}{false}
\newcommand{\twoCol}[2]{\ifthenelse{\boolean{twoColVersion}} {#1} {#2} }